\documentclass[journal,12pt,onecolumn,draftclsnofoot]{IEEEtran}

\usepackage{amsmath,graphicx}
\usepackage{epstopdf}
\usepackage{algorithm}
\usepackage{algorithmic}
\usepackage{array}
\usepackage{url}
\usepackage[colorlinks = true,
            linkcolor = blue,
            urlcolor  = blue,
            citecolor = blue,
            anchorcolor = blue]{hyperref}
\usepackage{multirow}
\usepackage{array}
\usepackage{textcomp}

\usepackage[table]{xcolor}
\usepackage{hhline}
\usepackage{pgf}
\usepackage{bm}
\usepackage{amssymb}

\newcommand*{\gray}{gray}
\newcommand*{\w}[1]{\multicolumn{1}{c}{#1}}
\newcommand*{\cm}[1]{%
    \pgfmathparse{#1<4?1:0}%
    \ifnum\pgfmathresult=0\relax\color{white}\fi
    \pgfmathparse{(13-#1)/13}
    \expandafter\cellcolor\expandafter[\expandafter\gray\expandafter]\expandafter{\pgfmathresult}%
    #1%
}
\newcommand*{\cn}[1]{%
    \pgfmathparse{#1<4?1:0}%
    \ifnum\pgfmathresult=0\relax\color{white}\fi
    \pgfmathparse{(8-#1)/8}
    \expandafter\cellcolor\expandafter[\expandafter\gray\expandafter]\expandafter{\pgfmathresult}%
    #1%
}
\DeclareMathOperator*{\minA}{min}
\DeclareMathOperator*{\maxA}{max}
\DeclareMathOperator*{\supA}{sup}

\begin{document}

\title{Marine Animal Classification with Correntropy Loss Based Multi-view Learning}

\author{Zheng~Cao,
        Shujian~Yu,
        Bing~Ouyang,~\IEEEmembership{Member,~IEEE},
        Fraser~Dalgleish,
        Anni~Vuorenkoski,
        Gabriel~Alsenas
        and~Jose~C.~Principe,~\IEEEmembership{Fellow,~IEEE}
\thanks{Z. Cao, S. Yu and J. C. Principe are with the Department of Electrical and Computer Engineering, University of Florida, Gainesville, FL, 32611 USA. e-mail: [zcao87, yusjlcy9011]@ufl.edu, principe@cnel.ufl.edu.}
\thanks{B. Ouyang, F. Dalgleish, A. Vuorenkoski and G. Alsenas are with Harbor Branch Oceanographic Institute, Florida Atlantic University, Fort Pierce, FL, 34946 USA. e-mail:[bouyang, fdalglei, adalglei, galsenas]@fau.edu.}
\thanks{Manuscript received January xx, xxxx; revised January xx, xxxx.}}

\markboth{Journal of \LaTeX\ Class Files,~Vol.~0, No.~0, January~xxxx}%
{Shell \MakeLowercase{\textit{et al.}}: Bare Demo of IEEEtran.cls for IEEE Journals}

\maketitle

\begin{abstract}
To analyze marine animals’ behavior, seasonal distribution and abundance, digital imagery can be acquired by visual or Lidar camera. Depending on the quantity and properties of acquired imagery, the animals are characterized as either features (shape, color, texture, etc.), or dissimilarity matrices derived from different shape analysis methods (shape context, internal distance shape context, etc.). For both cases, multi-view learning is critical in integrating more than one set of feature/dissimilarity matrix for higher classification accuracy.  This paper adopts correntropy loss as cost function in multi-view learning, which has favorable statistical properties for rejecting noise. For the case of features, the correntropy loss-based multi-view learning and its "entrywise" variation are developed based on the multi-view intact space learning algorithm. For the case of dissimilarity matrices, the robust Euclidean embedding algorithm is extended to its multi-view form with the correntropy loss function. Results from simulated data and real-world marine animal imagery show that the proposed algorithms can effectively enhance classification rate, as well as suppress noise under different noise conditions.
\end{abstract}


\IEEEpeerreviewmaketitle

\section{Introduction}

The study of marine animals’ behavior, seasonal distribution and abundance is vital for various environmental agencies, commercial fishermen and marine research institutes. To this end, extensive amount of digital imagery and video are acquired by underwater vehicles (AUVs) and remotely operated vehicles (ROVs). Color images are more intuitive to humans compared sonar signal, yet manual labeling of images is still a daunting task considering the sheer size of the data. An automated solution is thus preferred, which consists of two major steps $-$ detection and classification. This paper will focus on the latter problem only. There are a good number of works concentrating on feature extraction for colored marine imagery. Shape (Fourier descriptors ), color (normalized color histograms), and texture (Gabor filters and grey-level co-occurrence)\cite{Huang13} are among the most exploited features. Biological characteristics such as body part ratio \cite{Huang13} can distinguish between species as well. Nowadays, the rapid development in convolutional neural networks (CNN) has opened new possibility for accurate image representation, which has since benefited marine animal classification. To acquire CNN features of an image, one can either input the image to a CNN pre-trained by a large database (e.g. ImageNet \cite{Deng09}) which consists of images that are visually similar to the target image \cite{Cao15_2}, or train a new CNN with images homogenous to the target as in the example of plankton classification \cite{dieleman2015classifying}.  

\begin{figure}[t]
\includegraphics[width=0.7\linewidth]{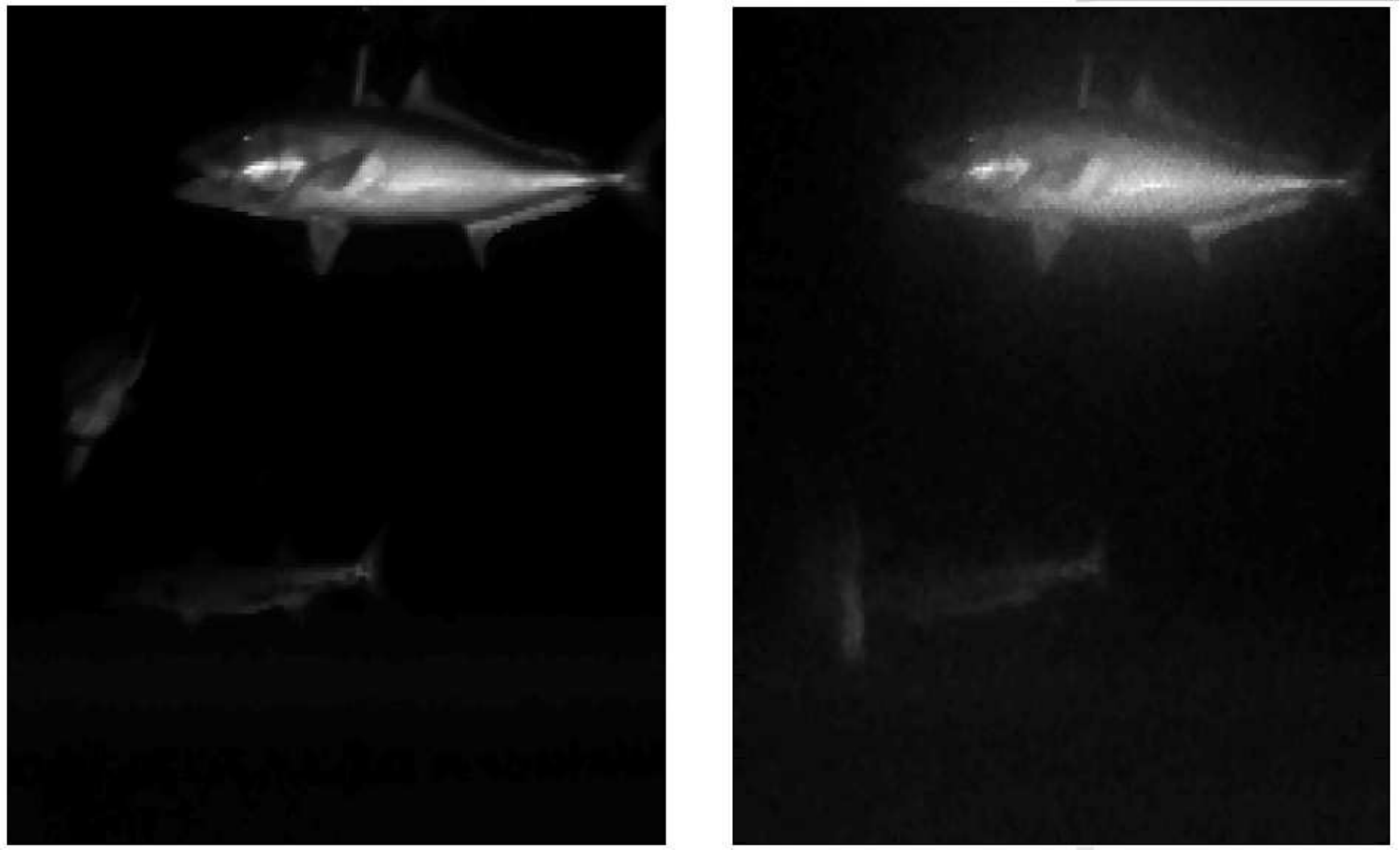}
\centering
\caption{Lidar image examples. Left image and right image are retrieved under clear water and turbid water (attenuation coefficient c=0.73) conditions respectively.}
\label{fig:joelidarimage}
\end{figure}

\begin{figure}[t]
\includegraphics[width=5in,height=2.5in]{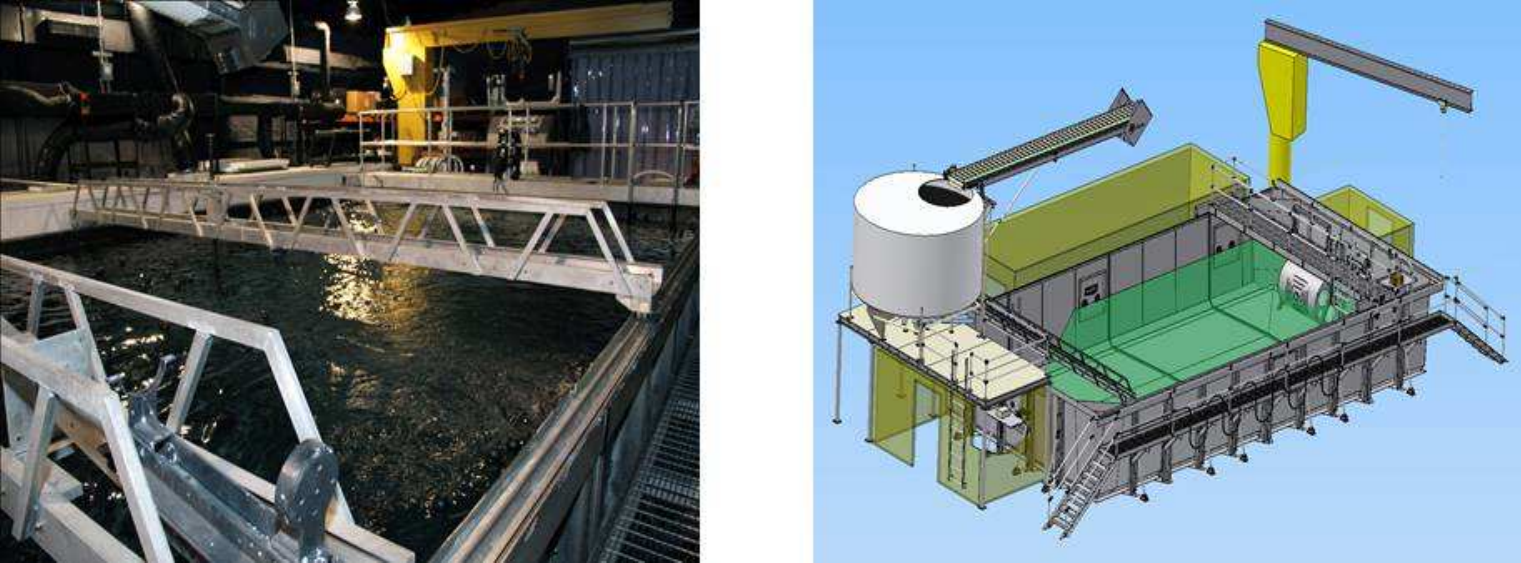}
\centering
\caption{HBOI Underwater Electro-Optics Testing Facility.}
\label{fig:testtank}
\end{figure}

Recently, the Harbor Branch Oceanographic Institute (HBOI) at Florida Atlantic University has developed a novel system called Unobtrusive Multistatic Serial Lidar Imager (UMSLI) to perform marine hydrokinetic site monitoring and marine animal classification \cite{ouyang2013visualization}\cite{Cao16_2}. Initial testing of the UMSLI system has been conducted inside a unique test tank facility (Figure \ref{fig:testtank}), which is capable of extensive testing of a variety of different electro-optical system configurations under a range of environmental conditions. Compared with optical camera imaging, underwater Lidar imaging has several advantages. Firstly, red laser illuminators are beyond the visible wavelength range of marine life, thus animals being monitored will not be affected. Meanwhile, optical camera requires significant amount of white light to illuminate low light areas and is more obtrusive to marine life. Second, unlike conventional camera whose focus is governed by the lens, Lidar imagery will remain in focus throughout the entire range, which gives it superior detection range. Higher SNR is also be achieved with Lidar due to the higher photon efficiency \cite{dalgleish2013extended}. Third, the transmitter in UMSLI system can operate in an adaptive mode, opting for either higher resolution or longer range of detection. A Lidar image of a fish captured by UMSLI is typically 2-D grayscale integrated from the 3-D point cloud Lidar return. Figure \ref{fig:joelidarimage} shows example Lidar images retrieved from the test tank at Harbor Branch Oceanographic Institute (HBOI). Using Lidar imagery for marine animal classification, however, is not without its own issues. As UMSLI is the first attempt to identify individual marine animal using Lidar imagery, there are virtually no existing database online with similar content. Given that the amount of data obtained from initial UMSLI deployment is also small, training a proper convolutional neural network becomes problematic with insufficient data. In addition, most of the valuable information revealed from the obtained Lidar imagery in the initial experimental dataset seems to be the shape of the animal. Using directly the pixels as feature is not recommended because of orientation variations among shapes, while traditional shape features such as Zernike \cite{Ouyang07} or Hu's moments have relatively weak description ability. There has been an attempt \cite{bai2014shape} to apply bag of words for quantifying a shape's feature, but the resulting "shape vocabulary" feature will be very long and highly redundant, because the common space in which all shapes reside can have very high dimensions. Instead of feature extraction, most of the existing shape classification or recognition literatures adopt a "pairwise" comparison strategy, creating directly a matrix with each entry representing the similarity/dissimilarity between a pair of shapes. As such, a "descriptor" rather than a "feature" will be enough to represent a shape, which is usually much simpler and intuitive. Commonly used descriptors include shape context (SC) \cite{Belongie02}, internal distance shape context  (IDSC) \cite{Ling07}, the triangle descriptor \cite{alajlan2008geometry} and height functions \cite{wang2012shape}. With a proper dissimilarity measure (Chi square or earth mover distance \cite{ling2007efficient}) applied on pairs of descriptors, these methods can usually achieve satisfying classification results using k-nearest neighbor classifier, especially on data with very few training samples. This similarity-matrix based approach is not limited to shape descriptors only. For instance, Gaussian mixture models \cite{kampa2011closed}\cite{liu2012shape} have been applied in hand gesture recognition, while spectral estimation \cite{sadeghian2015automatic} method has been used in an automobile recognition task. 

For both feature based and dissimilarity matrix-based marine animal classification, utilizing information from multiple sources (different features/descriptors or different view angles) will lead to more comprehensive description of objects, and thus improve classification accuracy \cite{Cao15_2}\cite{Cao16_2}. Multi-view learning \cite{xu2013survey} is a group of methods that introduces one function to model one particular view of the data, then jointly optimizes all the functions to improve the learning performance. The goal of this paper is to develop suitable multi-view learning algorithms for both data formats, where a "view" would be respectively a feature set or a dissimilarity matrix. There are two major considerations when designing a multi-view learning algorithm. Firstly, the algorithm should simultaneously accommodate dimensional reduction. because it is a necessary preprocessing step that makes classification faster, and also more accurate when the data is small or has a low-dimensional structure \cite{lacoste2009disclda}\cite{wang2014role}. CNN features usually are high dimensional (4096 for DeCAF \cite{Donahue14}) and have considerable redundancy \cite{Cao15}, so it is natural to apply dimensionality reduction. With this principle in mind, one can choose the desired multi-view learning framework from various options.  For features, the most notable categories of multi-view learning algorithms are co-training \cite{blum1998combining}, multiple kernel learning (MKL) \cite{gonen2011multiple} and subspace learning \cite{xu2013survey}. Only subspace learning involves dimensionality reduction. Examples include multiple spectral embedding \cite{xia2010multiview}, multi-view non-negative matrix factorization \cite{ou2016multi} and multi-view intact space learning \cite{xu2015multi}. For dissimilarity matrices, there are very few well-established multi-view learning algorithms. It should be noted that MKL cannot be applied to dissimilarity matrices, as they are not guaranteed to be positive semi-definite and hence are not valid kernel matrices. Co-transduction \cite{bai2010co} borrows ideas from co-training and can be viewed as a multi-view approach, yet it lacks a dimensionality reduction mechanism. It may be a good idea to firstly find a “base” method that performs dimensionality reduction for a single dissimilarity matrix, then extend it to its multi-view version. Potential candidates include principal component analysis (PCA), multidimensional scaling (MDS) and auto-encoder \cite{wang2013feature}. Before carrying out any of these algorithms, it is necessary to enforce the dissimilarity matrix to be a proper distance matrix. The robust Euclidean embedding (REE) \cite{cayton2006robust} is an algorithm based on classical MDS, which also regulates the dissimilarity matrix by enforcing Euclidean distance. Meanwhile, there is actually one algorithm in the literature with the name multi-view MDS (MV-MDS) \cite{Bai16}, but it is based on non-classical MDS and is compatible only with the $L_2$ cost function, unlike REE.

Another important consideration in designing multi-view learning algorithms is the choice of cost functions, which has not been studied in detail in previous works. In a dataset, the views given may contain irregularities (noise) of different types and magnitudes. The mean square error (MSE) is the most widely-used cost function, yet its performance is suboptimal for non-Gaussian noise. The correntropy \cite{Liu07} as a non-linear, local similarity measure that is robust to outliers has attracted researchers in recent years. Notable applications of correntropy include adaptive filtering \cite{zhao2011kernel}, classification \cite{pokharel2012kernel}, face recognition \cite{he2011maximum} and robust autoencoder \cite{qi2014robust}. More recently, the generalized correntropy \cite{chen2015generalized} is proposed and successfully applied to adaptive filtering. It is more versatile than correntropy, as changes in the shape parameter can lead to the suppression of different types of noise.

In this paper, correntropy loss based multi-view learning algorithms will be developed for both features and dissimilarity matrices. For features, the multi-view intact space learning (MISL) \cite{xu2015multi} is employed as the base method for two related correntropy-based multi-view learning algorithms. For dissimilarity matrices, the base method will be REE. As REE is itself single view only, its direct multi-view version will be proposed along with a correntropy-based method. The rest of this paper is organized as follows: Section II reviews the MISL and REE algorithms as well as the concept of generalized correntropy. Section III derives the algorithms. Section IV shows experiment results for both simulated data and real-world marine animal data. Section V concludes the paper.

\section{Background}

\subsection{The MISL algorithm}
The MISL \cite{xu2015multi} algorithm aims at learning a low-dimensional latent intact subspace from two or more different views. One advantage of MISL is that it does not need the assumption that each view needs to be sufficient; as long as enough views are given, the learned view will be "intact", or fully able to describe the object. MISL uses the Cauchy loss  
\begin{align}
J_{cauchy}(e)&=\log (1+\frac{e^2}{c^2})
\label{eqn:cauchy}
\end{align}
to minimize the reconstruction error over the latent intact space:
\begin{align}
\begin{split}
\minA_{\textbf{x},\textbf{W}}&\frac{1}{MN}\sum_{v=1}^M\sum_{i=1}^N{\log}(1+\frac{||\textbf{z}_i^{(v)}-\textbf{W}_{(v)}\textbf{x}_i||_2^2}{c^2})\\ &+C_1\sum_{v=1}^m||\textbf{W}_{(v)}||_F^2+C_2\sum_{i=1}^n||\textbf{x}_i||_2^2
\end{split}
\label{eqn:misl}
\end{align}
In (\ref{eqn:cauchy}), $e$ refers to error and $c$ is a user-defined shape parameter. In (\ref{eqn:misl}), vectors $\textbf{z}_i^{(v)}$ and $\textbf{x}_i$ stand for, respectively, the $d_{(v)}$*1 feature vector from the $v^{th}$ view and the $d$*1 "common view" feature vector to be learned at the $i^{th}$ instance, while $\textbf{W}_{(v)}$ is the $v^{th}$ transformation matrix with dimensions $d_{(v)}$*$d$. There are $M$ views and $N$ instances in total.  The iteratively ieweight residuals (IRR) technique is used to find a solution for (\ref{eqn:misl}).  

\subsection{The REE algorithm}
The classical multidimensional scaling (cMDS) \cite{wickelmaier2003introduction} seeks to find the low-dimensional data representation $\textbf{X}$, whose associated Euclidean distance matrix $\textbf{D}$ approximates the given dissimilarity matrix $\mathbf{\Delta}$. Both $\textbf{D}$ and $\mathbf{\Delta}$ are $N$ by $N$ ($N$ is the data size), while $\textbf{X}$ is $N$ by $k$ ($k<N$). The cMDS is processed in two stages. Firstly, the dissimilarity matrix is embedded to the Euclidean distance space, which solves the following optimization problem 
\begin{equation}
\minA_\textbf{D} ||\textbf{H}\mathbf{\Delta} \textbf{H}-\textbf{HDH}||_2^2
\label{eqn:cmds}
\end{equation}
where $\textbf{H}=\textbf{I}-\frac{1}{N}\textbf{1}\textbf{1}^T$is the centering matrix. Second, dimensionality of $\textbf{X}$ is reduced through PCA.

The robust Euclidean embedding (REE) \cite{cayton2006robust} algorithm states that the robustness of the Euclidean embedding process can be enhanced by two practices: (1) Replacing the $L_2$ norm with $L_1$ norm in the optimization function; (2) Rather than projecting the matrix $\textbf{B}=-\frac{1}{2}\textbf{H}\mathbf{\Delta} \textbf{H}$ to the positive semidefinite cone, REE takes a direct approach by projecting $\mathbf{\Delta}$ onto the Euclidean distance matrix. Therefore, REE seeks to solve the optimization problem
\begin{equation}
\minA_\textbf{D} W_{ij}|\Delta_{ij}-D_{ij}|
\label{eqn:REE}
\end{equation}
The weighting matrix $\textbf{W}$ is usually set to all ones. The Gram matrix $\textbf{B}$ associated with $\textbf{D}$ should still meet the condition of being positive semidefinite. Since $\textbf{D}$ and $\textbf{B}$ are related by
\begin{equation}
D_{ij}=B_{ii}+B_{jj}-B_{ij}-B_{ji}
\label{eqn:DandB}
\end{equation}
(\ref{eqn:REE}) can be optimized by taking subgradient with respect to $\textbf{B}$, and constrain $\textbf{B}$ to be positive semidefinite at every iteration. 

\subsection{The generalized correntropy loss function}
\label{sec:gcloss}
\begin{figure}[t]
\includegraphics[width=0.8\linewidth]{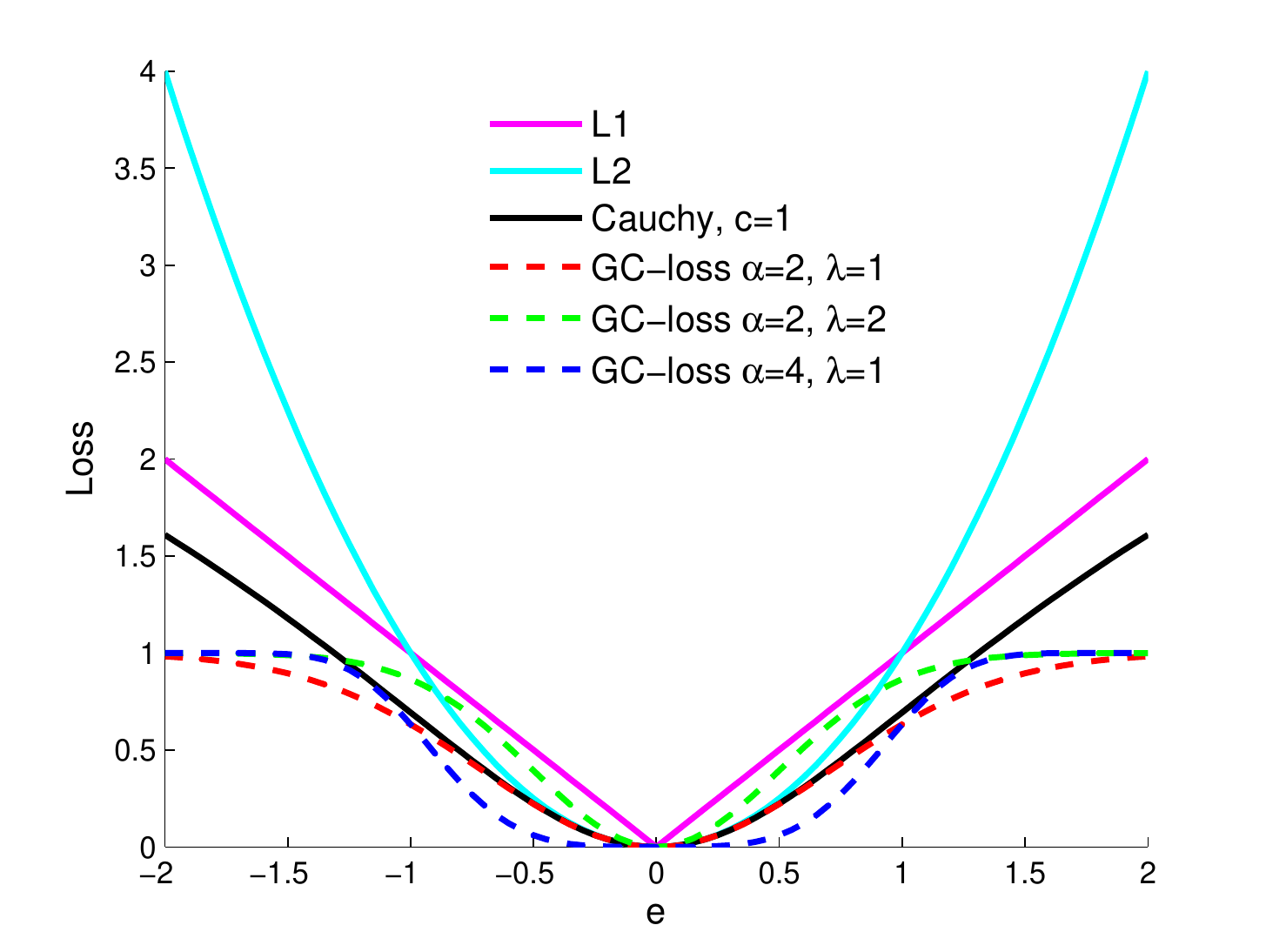}
\centering
\caption{Comparison of different loss functions. The normalizing constant $\gamma$ is ignored for GC-loss.}
\label{fig:gcloss}
\end{figure}

The generalized correntropy is a similarity measure between two random variables $X$ and $Y$ \cite{chen2015generalized}
\begin{equation}
V(X,Y)=\textbf{E}[G_{\alpha,\beta}(X-Y)]
\label{eqn:GC}
\end{equation}
In (\ref{eqn:GC}), $G_{\alpha,\beta}(\cdot)$ is called the generalized Gaussian density (GGD) function
\begin{equation}
G_{\alpha,\beta}(e)=\frac{\alpha}{2\beta\Gamma(\frac{1}{\alpha})}\exp{\Big{(}-|\frac{e}{\beta}|^{\alpha}\Big{)}}=\gamma_{\alpha,\beta} \exp{\big{(}-\lambda|e|^{\alpha}\big{)}}
\label{eqn:GGD}
\end{equation}
where $\alpha >0$ is the shape parameter, $\beta >0$ is the bandwidth parameter, $\lambda=1/\beta^{\alpha}$ is the kernel parameter and $\gamma_{\alpha,\beta}=\frac{\alpha}{2\beta\Gamma(1/\alpha)}$ is the normalizing constant. When $\alpha=2$, the GGD becomes a Gaussian kernel $\kappa(e)=\frac{\sqrt{\lambda}}{\sqrt{\pi}}\exp{(-\lambda e^2)}=\frac{1}{\sigma\sqrt{2\pi}}\exp{(-\frac{e^2}{2\sigma^2})}$, which turns $V(X,Y)$ into correntropy - a well known specific case of generalized correntropy. The generalized correntropy has the advantages of being smooth, positive and bounded. It also involves higher order absolute moments of the error variable: 
\begin{equation}
V(X,Y)=\gamma_{\alpha,\beta} \sum_{n=0}^{\infty} \frac{(-\lambda)^n}{n!} E\textbf{•}\big{[}|X-Y|^{\alpha n}\big{]}
\label{eqn:higher}
\end{equation}
In classification tasks, a loss function based on correntropy called correntropy loss is used \cite{syed2014optimization}. The generalized correntropy loss (GC-loss) as a function of error takes the following form\begin{align}
J_{GC-loss}(e)&=G_{\alpha, \beta}(0)-G_{\alpha, \beta}(e)\nonumber \\&= \gamma_{\alpha,\beta} \big{(}1-\exp{(-\lambda|e|^{\alpha})}\big{)}
\label{eqn:gcloss}
\end{align}
Figure \ref{fig:gcloss} shows several loss functions. Clearly, neither $L_1$ nor Cauchy loss is bounded, unlike GC-loss. In fact, Cauchy loss would resemble GC-loss with $\alpha=2$ (correntropy loss) when error $e$ is small, according to their Taylor expansion:
\begin{equation*}
\begin{split}
\log(1+\frac{e^2}{c^2})=\frac{e^2}{c^2}-\frac{e^4}{2c^4}+\frac{e^6}{3c^6}...\\
1-\exp{(-\lambda|e|^2)}=\lambda e^2-\frac{\lambda ^2 e^4}{2!}+\frac{\lambda^3 e^6}{3!}...
\label{eqn:Taylor}
\end{split}
\end{equation*}
Given that $\lambda = 1/c^2$, the first two dominant terms are the same for both loss functions. However, Cauchy will suffer from its unboundedness when $e$ is large. GC-loss has the interesting property of behaving like different norms for different values of $e$ \cite{chen2015generalized}. When $e$ is very small, GC-loss acts like L$\alpha$ norm. As $e$ increases, GC-loss moves gradually towards L0 norm. Therefore, different choices for the shape parameter $\alpha$ are beneficial for different types of noise. Smaller $\alpha$ is better when distribution of noise is heavy-tailed (e.g. Laplace, $\alpha$ stable), while larger $\alpha$ is better for light tailed noise (e.g. uniform, binary).

\section{The correntropy loss based multi-view learning algorithm}

\subsection{Algorithm for features}
\label{sec:c-mv-feat}

According to previous analysis, the cost function for correntropy-loss based multi-view learning (C-MV) can be written as
\begin{align}
\begin{split}
\minA_{\textbf{x},\textbf{W}}R_1(\textbf{x},\textbf{W})=&\frac{1}{MN}\sum_{v=1}^M\sum_{i=1}^N{\gamma}_{\sigma}[1-{\kappa}_{\sigma}(||\textbf{z}_i^{(v)}-\textbf{W}_{(v)}\textbf{x}_i||_2)]\\&+C_{10}\sum_{v=1}^m||\textbf{W}_{(v)}||_F^2+C_{20}\sum_{i=1}^n||\textbf{x}_i||_2^2 
\end{split}
\label{eqn:cmvfeat_ori}
\end{align}
where $\kappa_{\sigma}$ is a Gaussian kernel. Expression (\ref{eqn:cmvfeat_ori}) can be rewritten in the more compact form
\begin{align}
\begin{split}
\maxA_{\textbf{x},\textbf{W}}R_2(\textbf{x},\textbf{W})=&\sum_{v=1}^M\sum_{i=1}^N{\exp}(-\frac{||\textbf{z}_i^{(v)}-\textbf{W}_{(v)}\textbf{x}_i||_2^2}{2\sigma^2})\\ &-C_1\sum_{v=1}^m||\textbf{W}_{(v)}||_F^2-C_2\sum_{i=1}^n||\textbf{x}_i||_2^2
\end{split}
\label{eqn:cmvfeat}
\end{align}
In (\ref{eqn:cmvfeat}), variables $\textbf{z}_i^{(v)}$, $\textbf{x}_i$, $\textbf{W}_{(v)}$ and parameters $d_{(v)}$, $C_1$, $C_2$ have the same connotation as in (\ref{eqn:misl}). The main goal is still to solve for the "common view" $\textbf{x}_i$. Kernel size $\sigma$ is usually set equal to or smaller than 1, provided that $z^{(v)}$ is normalized: $\textbf{z}^{(v)}:=\frac{\textbf{z}^{(v)}}{\sum_i^N||\textbf{z}_i^{(v)}||_2^2/N}$. Regularization parameters $C_1$ and $C_2$ can be chosen using cross validation.

To solve (\ref{eqn:cmvfeat}) with HQ optimization, a convex function is defined as $g(a)=-a\ln(a)+a$, where $a<0$. The conjugate function $g^*(b)$ of $g(a)$ is then \cite{boyd2004convex} 
\begin{equation}
g^*(b)=\sup_{a<0}(ba-g(a)=\exp{(-b)}
\label{eqn:gb}
\end{equation}
In (\ref{eqn:gb}), the supremum is achieved when $a=\exp{(-b)}<0$. Define $b_i^{(v)}=+\frac{||\textbf{z}_i^v-\textbf{W}_v\textbf{x}_i||_2^2}{2\sigma^2}$. It follows that 
\begin{align*}
&\;\;\;\;g^*(\frac{||\textbf{z}_i^v-\textbf{W}_v\textbf{x}_i||_2^2}{2\sigma^2})\\ \nonumber
&=\sup_{a_i^{(v)}<0}\Big{\lbrace}\frac{||\textbf{z}_i^v-\textbf{W}_v\textbf{x}_i||_2^2}{2\sigma^2}a_i^{(v)}-g(a_i^{(v)})\Big{\rbrace}\\ \nonumber
&=\exp{(-\frac{||\textbf{z}_i^v-\textbf{W}_v\textbf{x}_i||_2^2}{2\sigma^2})}
\end{align*}
where $a_i^{(v)}=-\exp{(-\frac{||\textbf{z}_i^v-\textbf{W}_v\textbf{x}_i||_2^2}{2\sigma^2})}<0$. Therefore,
\begin{equation}
\begin{split}
\;\;\;\;R_2(\textbf{x},\textbf{W})=&\sum_{v=1}^M\sum_{i=1}^N\supA_{a_i^v<0}\Big{\lbrace}\frac{||\textbf{z}_i^v-\textbf{W}_v\textbf{x}_i||_2^2}{2\sigma^2}a_i^v-g(a_i^v)\Big{\rbrace}\\ 
\;\;\;\;&-C_1\sum_{v=1}^M||\textbf{W}_v||_F^2-C_2\sum_{i=1}^N||\textbf{x}_i||_2^2\\ 
=&\supA_{\textbf{A}\prec0}\Big{\lbrace}\sum_{v=1}^M\sum_{i=1}^N\big{[}\frac{||\textbf{z}_i^v-\textbf{W}_v\textbf{x}_i||_2^2}{2\sigma^2}a_i^v-g(a_i^v)\big{]}\\ 
\;\;\;\;&-C_1\sum_{v=1}^M||\textbf{W}_v||_F^2-C_2\sum_{i=1}^N||\textbf{x}_i||_2^2\Big{\rbrace}
\end{split}
\label{eqn:cmvfeat3_0}
\end{equation}
With (\ref{eqn:cmvfeat3_0}), (\ref{eqn:cmvfeat}) is equivalent to
\begin{equation}
\begin{split}
\maxA_{\textbf{x},\textbf{W},\textbf{A}\prec0}R_3(\textbf{x},\textbf{W},\textbf{A})=&\sum_{v=1}^M\sum_{i=1}^N\big{[}\frac{||\textbf{z}_i^v-\textbf{W}_v\textbf{x}_i||_2^2}{2\sigma^2}a_i^v-g(a_i^v)\big{]}\\ 
\;\;\;\;&-C_1\sum_{v=1}^M||\textbf{W}_v||_F^2-C_2\sum_{i=1}^N||\textbf{x}_i||_2^2
\label{eqn:cmvfeat3}
\end{split}
\end{equation}
where $\textbf{A}_{(v,i)}=a_i^{(v)}<0$. A double-loop alternating optimization scheme can be applied to optimize (\ref{eqn:cmvfeat3}). In the outer loop, the alternating optimization is between \textbf{A} and \{\textbf{x}, \textbf{W}\}. Given $x_i$ and $\textbf{W}_{(v)}$ for all $i$ and $v$, (\ref{eqn:cmvfeat3}) becomes equivalent to 
\begin{equation}
\maxA_{\textbf{A}}\sum_{v=1}^M\sum_{i=1}^N\big{[}\frac{||\textbf{z}_i^v-\textbf{W}_v\textbf{x}_i||_2^2}{2\sigma^2}a_i^v-g(a_i^v)\big{]}
\label{eqn:awx_a}
\end{equation}
whose analytical solution is
\begin{equation}
\begin{split}
a_i^{(v)}=-{\exp}(-\frac{||\textbf{z}_i^{(v)}-\textbf{W}_{(v)}\textbf{x}_i||_2^2}{2\sigma^2}) \\
\end{split}
\label{eqn:sola}
\end{equation}
When $a_i^{(v)}$ is given, (\ref{eqn:cmvfeat3}) is equivalent to 
\begin{equation}
\begin{split}
\;\;\;\;\maxA_{\textbf{x},\textbf{W}}&\sum_{v=1}^M\sum_{i=1}^N\frac{a_i^{(v)}}{2\sigma^2}||\textbf{z}_i^{(v)}-\textbf{W}_{(v)}\textbf{x}_i||_2^2\\ 
\;\;\;\;&-C_1\sum_{v=1}^M||\textbf{W}_{(v)}||_F^2-C_2\sum_{i=1}^N||\textbf{x}_i||_2^2\\ 
\Rightarrow\minA_{\textbf{x},\textbf{W}}&\sum_{v=1}^M\sum_{i=1}^N(-a_i^{(v)})||\textbf{z}_i^{(v)}-\textbf{W}_{(v)}\textbf{x}_i||_2^2\\ 
\;\;\;\;&+C_1\sum_{v=1}^M||\textbf{W}_{(v)}||_F^2+C_2\sum_{i=1}^N||\textbf{x}_i||_2^2
\end{split}
\label{eqn:awx_wx}
\end{equation}
In the latter equation $C_1:=C_1*2\sigma^2$, $C_2:=C_2*2\sigma^2$. In the inner loop, the alternating optimization is between \textbf{W} and \textbf{x}. Given $\textbf{W}_{(v)}$, (\ref{eqn:awx_wx}) becomes $N$ independent problems:
\begin{equation}
\minA_{\textbf{x}_i}\sum_{v=1}^M(-a_i^{(v)})||\textbf{z}_i^{(v)}-\textbf{W}_{(v)}\textbf{x}_i||_2^2+C_2||\textbf{x}_i||_2^2
\label{eqn:wx_x}
\end{equation}
For (\ref{eqn:wx_x}), taking the derivative with respect to $\textbf{x}_i$ and equating it to zero one will get
\begin{equation}
\textbf{x}_i=(\sum_{v=1}^Ma_i^v\textbf{W}_v^T\textbf{W}_{(v)}-C_2\textbf{I})^{-1}(\sum_{v=1}^Ma_i^{(v)}\textbf{W}_{(v)}^T\textbf{z}_i^{(v)}) \\
\label{eqn:solx}
\end{equation}
When $\textbf{x}_i$ is given, (\ref{eqn:awx_wx}) becomes $M$ independent problems:
\begin{equation}
\minA_{\textbf{W}^{(v)}}\sum_{i=1}^N(-a_i^{(v)})||\textbf{z}_i^{(v)}-\textbf{W}_{(v)}\textbf{x}_i||_2^2+C_1||\textbf{W}_{(v)}||_F^2
\label{eqn:wx_w}
\end{equation}
Let the derivative with respect to $\textbf{W}^{(v)}$ equal zero and solve for $\textbf{W}^{(v)}$ one will get
\begin{equation}
\textbf{W}_{(v)}=(\sum_{i=1}^Na_i^{(v)}\textbf{z}_i^{(v)}\textbf{x}_i^T)(\sum_{i=1}^Na_i^{(v)}\textbf{x}_i\textbf{x}_i^T-C_1\textbf{I})^{-1}
\label{eqn:solw}
\end{equation}

Alternatively to C-MV, this paper will propose another multi-view learning algorithm for features, called correntropy-loss entrywise multi-view learning for features (Ce-MV) to mitigate some of the deficiencies of (\ref{eqn:cmvfeat}). Its optimization goal is
\begin{align}
\begin{split}
\maxA_{\textbf{x},\textbf{W}}&\sum_{v=1}^M\frac{1}{d_{(v)}}\sum_{i=1}^N\sum_{j=1}^{d_{(v)}}{\exp}(-\frac{(z_{ij}^{(v)}-\textbf{W}_j^{(v)}\textbf{x}_i)^2}{2\sigma^2})\\ &-C_1\sum_{v=1}^m\sum_{j=1}^{d_{(v)}}||\textbf{W}_j^{(v)}||_F^2-C_2\sum_{i=1}^n||\textbf{x}_i||_2^2
\end{split}
\label{eqn:entrycmvfeat}
\end{align}
Notations of variables and parameters in (\ref{eqn:entrycmvfeat}) are the same as in (\ref{eqn:cmvfeat}). The coefficient $\frac{1}{d_{(v)}}$ ensures that features with higher dimensions are given the same weight as lower dimensional feature set. The main difference between (\ref{eqn:entrycmvfeat}) and (\ref{eqn:cmvfeat}) is that Ce-MV maximizes the overall correntropy over every individual entry $z_{ij}$ of the input feature, not just every feature vector $\textbf{z}_i$. The potential advantage is that when certain entries of $\textbf{z}_i$ have abnormal values caused by noise or low quality features, the whole $\textbf{z}_i$ will be be affected as a result of the $L_2$ norm used inside the correntropy function in (\ref{eqn:cmvfeat}). On the other hand, abnormal values in $\textbf{z}_i$ will restrict its effect to itself only as in (\ref{eqn:entrycmvfeat}), making the "good" features contribute more effectively. A good choice for kernel size $\sigma$ for Ce-MV should be much smaller than that for C-MV, preferrably rescaled to $1/\sqrt{d_{(v)}}$ of the original.

The optimization for Ce-MV can be done following the same HQ and alternating optimization technique. Expression (\ref{eqn:entrycmvfeat}) is equivalent to 
\begin{equation}
\begin{split}
\Rightarrow\maxA_{\textbf{x},\textbf{W},\textbf{A}}&\supA_{\textbf{A}\prec0}\Big{\lbrace}\sum_{v=1}^M\frac{1}{d_v}\sum_{i=1}^N\sum_{j=1}^{d_{(v)}}\big{[}\frac{(z_{ij}^{(v)}-\textbf{W}_j^{(v)}\textbf{x}_i)^2}{2\sigma_{(v)}^2}a_{ij}^{(v)}-g(a_{ij}^v)\big{]}\\ 
\;\;\;\;&-C_1\sum_{v=1}^m\sum_{j=1}^{d_{(v)}}||\textbf{W}_j^{(v)}||_F^2-C_2\sum_{i=1}^n||\textbf{x}_i||_2^2\Big{\rbrace}
\end{split}
\label{eqn:entrycmvfeat2}
\end{equation}
Solution for $a_{ij}^{(v)}$ in the outer loop of alternating optimization is 
\begin{equation}
a_{ij}^{(v)}=-{\exp}(-\frac{(z_{ij}^{(v)}-\textbf{W}_j^{(v)}\textbf{x}_i)^2}{2\sigma_{(v)}^2}) \\
\label{eqn:entrysola}
\end{equation}
In the inner loop, solutions for $\textbf{x}_i$ and $\textbf{W}_j^{(v)}$ are
\begin{equation}
\begin{split}
\textbf{x}_i=&\big{(}\sum_{v=1}^M\frac{1}{d_v}\sum_{j=1}^{d_{(v)}}a_{ij}^{(v)}(\textbf{W}_j^{(v)})^T\textbf{W}_j^{(v)}-C_2\textbf{I}\big{)}^{-1}\\&\big{(}\sum_{v=1}^M\frac{1}{d_v}\sum_{j=1}^{d_{(v)}}a_{ij}^{(v)}z_{ij}^{(v)})(\textbf{W}_j^{(v)})^T \big{)}
\label{eqn:entrysolx}
\end{split}
\end{equation}
\begin{equation}
\textbf{W}_j^{(v)}=(\sum_{i=1}^Na_{ij}^{(v)}z_{ij}^{(v)}\textbf{x}_i^T)(\sum_{i=1}^Na_{ij}^{(v)}\textbf{x}_i\textbf{x}_i^T-C_1\textbf{I})^{-1}
\label{eqn:entrysolw}
\end{equation}
Note that the intermediate variables $a_i^{(v)}$ for C-MV and $a_{ij}^{(v)}$ for Ce-MV, are reflective of the importance of an instance $\textbf{z}_i$ when the algorithm converges. Additionally, $a_{ij}^{(v)}$ can reveal the importance of particular features within the instance, which cannot be said for $a_i^{(v)}$. 

\textit{Convergence analysis.} It can be proved that for C-MV, the sequence $R_3(\textbf{x}^k,\textbf{W}^k,\textbf{A}^k)$ (k=1,2,... stands for the number of outer iteration) in \ref{eqn:cmvfeat3} converges: from (\ref{eqn:cmvfeat}), (\ref{eqn:cmvfeat3_0}) and (\ref{eqn:cmvfeat3}), it is clear that $R_3(\textbf{x},\textbf{W},\textbf{A})\leq R_2(\textbf{x},\textbf{W}) \leq MN$, which means $R_3(\textbf{x},\textbf{W},\textbf{A})$ is upper bounded. Then, from (\ref{eqn:awx_a}) and (\ref{eqn:awx_wx}), it can be concluded that $R_3(\textbf{x}^k,\textbf{W}^k,\textbf{A}^k)\leq R_3(\textbf{x}^k,\textbf{W}^k,\textbf{A}^{k+1}) \leq R_3(\textbf{x}^{k+1},\textbf{W}^{k+1},\textbf{A}^{k+1})$, i.e. $R_3(\textbf{x}^k,\textbf{W}^k,\textbf{A}^k)$ is non-decreasing. Therefore, the sequence $R_3(\textbf{x}^k,\textbf{W}^k,\textbf{A}^k)$ (k=1,2,...) converges. By the same token, convergence of Ce-MV can be proven as well.

The C-MV and Ce-MV algorithms are summarized in Algorithm \ref{alg:C-MV}. Disadvantage of Ce-MV is that it is much slower, especially when the dimensions of original features $d_{(v)}$ are high, because computing $d_{(v)}$ values of $a_{ij}^{(v)}$ and $\textbf{W}_j^{(v)}$ are more time consuming than computing a single value of $a_{i}^{(v)}$ and $\textbf{W}^{(v)}$.

 \begin{algorithm}[H]
 \caption{Algorithm for C-MV / Ce-MV}
 \label{alg:C-MV}
 \begin{algorithmic}[1]
 \renewcommand{\algorithmicrequire}{\textbf{Input:}}
 \REQUIRE $\textbf{z}^{(v)}$
 \renewcommand{\algorithmicrequire}{\textbf{Initialization:}}
 \REQUIRE $\textbf{W}^{(v)}$, $\textbf{x}$, $\sigma$, $C_1$, $C_2$
 \FOR {$k$ = 1 to maximum outer iteration}
 \STATE update $a_{i}^{(v)}$ as in (\ref{eqn:sola}) /  $a_{ij}^{(v)}$ as in (\ref{eqn:entrysola}) for all subscripts 
 \FOR {$k_i$ = 1 to maximum inner iteration}
 \STATE update $\textbf{x}_i$ as in (\ref{eqn:solx}) / (\ref{eqn:entrysolx}) for all $i$
 \STATE update $\textbf{W}^{(v)}$ as in (\ref{eqn:solw}) /  $\textbf{W}^{(v)}_j$ as in (\ref{eqn:entrysolw}) for all subscripts
 \ENDFOR
 \ENDFOR
 \RETURN $\textbf{x}$
 \end{algorithmic}
 \end{algorithm}

\subsection{Algorithm for dissimilarity matrices}
\label{sec:c-mv-ree}
Firstly, the REE algorithm can be naturally extended to its own multi-view learning version, which will be called multi-view REE (MV-REE) in this paper. Suppose the multiple views consist of $M$ $N*N$ real-valued squared dissimilarity matrices $\mathbf{\Delta}^{(1)}, ..., \mathbf{\Delta}^{(M)}$, with $(\mathbf{\Delta}^{(v)})^{T}=\mathbf{\Delta}^{(v)}$ and $\textrm{diag} (\mathbf{\Delta}^{(v)})=\textbf{0}$. Meanwhile, a "common view" for all $M$ views is the squared Euclidean distance matrix $\textbf{D}$. A cost function can be written  in a similar manner to (\ref{eqn:REE}), aiming at the minimization of the overall $L_1$ distance between all views and the common view:
\begin{equation}
\min_{\textbf{B}} f_0, f_0=\min\sum_{v=1}^M \sum_{ij} W_{ij}^{(v)}|\Delta^{(v)}_{ij}-D_{ij}|
\label{eqn:MV-REE}
\end{equation}
One can optimize (\ref{eqn:MV-REE}) with the same subgradient approach. A subgradient for $f_0$ with respect to the Gram matrix \textbf{B} is
\begin{equation}
\big[g_0(\textbf{B})\big]_{ij}=
\begin{cases}
    -\sum_{v=1}^M  W_{ij}^{(v)}\textrm{sign}(D_{ij}-\Delta^{(v)}_{ij}),& i\neq j,  \\
    \sum_{v=1}^M  \sum_{k=1}^N W_{ik}^{(v)}\textrm{sign}(D_{ik}-\Delta^{(v)}_{ik}),& i=j,
\end{cases}
\;\;\;\;\; 
\label{eqn:MV-REEsol}
\end{equation}
\noindent where $\textbf{B}$ is associated with $\textbf{D}$ as in (\ref{eqn:DandB}).

The proposed correntropy loss based multi-view robust Euclidean embedding (C-MV-REE) replaces the $L_1$ cost function in (\ref{eqn:REE}) with the correntropy loss:
\begin{equation}
\max_{\textbf{B}} f, f=\sum_{v=1}^M \sum_{ij} W_{ij}^{(v)}\exp\Big(-\frac{(\Delta^{(v)}_{ij}-D_{ij})^2}{2\sigma^2}\Big)
\label{eqn:C-MV-REE}
\end{equation}
\noindent Kernel size $\sigma$ can be set equal to or smaller than the median values of the dissimilarity matrices. The goal is to optimize (\ref{eqn:C-MV-REE}) with respect to $\textbf{B}$. Unlike REE which has a non-differentiable cost function, expression (\ref{eqn:C-MV-REE}) is differentiable and a gradient ascent optimization approach can be thus adopted. According to matrix calculus, the derivative of $f$ with respect to $\textbf{B}$ is
\begin{equation}
\frac{\partial f}{\partial B_{ij}}=\textrm{trace} (\frac{\partial f}{\partial \textbf{D}} \frac{\partial \textbf{D}}{\partial B_{ij}})
\label{eqn:pfpb}
\end{equation}
where 
\begin{equation}
\big[\frac{\partial f}{\partial \textbf{D}}\big]_{ij}=\sum_{v=1}^M  W_{ij}^{(v)}\exp\Big(-\frac{(\Delta^{(v)}_{ij}-D_{ij})^2}{2\sigma^2}\Big)\cdot\frac{\Delta^{(v)}_{ij}-D_{ij}}{\sigma^2}
\label{eqn:pfpd}
\end{equation}
and
\begin{align}
\nonumber
\big[\frac{\partial \textbf{D}}{\partial B_{ij}}\big]_{i'j'}=
\begin{cases}
    -1,& (i'=i,j'=j) \text{ or }\\
     &(i'=j,j'=i), \\
    0,& \text{others},
\end{cases}
\;\;\;\;\; \text{if } i\neq j
\\ 
\big[\frac{\partial \textbf{D}}{\partial B_{ij}}\big]_{i'j'}=
\begin{cases}
    2,& i'=i \text{ and } j'=i,\\
    1,& i'=i \text{ xor } j'=i,\\
    0,& \text{others},
\end{cases}
\;\;\;\;\; \text{if } i=j
\label{eqn:pdpb}
\end{align}
Therefore, 
\begin{align}
 \nonumber
\frac{\partial f}{\partial B_{ij}}=\sum_{v=1}^M  W_{ij}^{(v)}\exp\Big(-\frac{(\Delta^{(v)}_{ij}-D_{ij})^2}{2\sigma^2}\Big)\cdot\frac{D_{ij}-\Delta^{(v)}_{ij}}{\sigma^2} \\
\frac{\partial f}{\partial B_{ij}}=\sum_{v=1}^M  \sum_{k=1}^N W_{ik}^{(v)}\exp\Big(-\frac{(\Delta^{(v)}_{ik}-D_{ik})^2}{2\sigma^2}\Big)\cdot\frac{\Delta^{(v)}_{ik}-D_{ik}}{\sigma^2}
\label{eqn:C-MV-REEsol}
\end{align}
In (\ref{eqn:C-MV-REEsol}), the upper and lower equations correspond to the cases when $i\neq j$ and $i=j$, respectively. 

\textit{Convergence analysis.} The correntropy loss function $f$ in (\ref{eqn:C-MV-REE}) is an $MN^2$ dimensional pseudoconvex function. When gradient ascent/descent approach is used for optimization, global convergence is guaranteed provided that the step size is sufficiently small  \cite{syed2014optimization}. This will be illustrated by the convergence curves (Figure \ref{fig:converge} and Figure \ref{fig:converge_kimia}) in the experiments section.

The C-MV-REE algorithm is summarized in Algorithm \ref{alg:C-MV-REE}. One can obtain the MV-REE algorithm by simply replacing the term $\frac{\partial f}{\partial \textbf{B}}$ by $-g_0(\textbf{B})$. The algorithm returns the $N*N$ configuration matrix \textbf{X}, whose first $k$ columns that corresponds to $k$ dominant eigenvalues of \textbf{B} comprises a new $N*k$ configuration matrix \textbf{X}'. \textbf{X}' can be seen as the explicit representation of the original dataset, from which a new dissimilarity matrix can be calculated. One can also treat \textbf{X}' as feature vectors, opening up the possibility of using classifiers other than kNN (such as SVM). 

Computational time of C-MV-REE is at the same order of magnitude as the base method REE. This is because C-MV-REE differs from REE at step 2 only in Algorithm \ref{alg:C-MV-REE}. It takes $M$ times as much time for MV-REE to run step 2 compared to REE. Meanwhile, the running time of step 2 for C-MV-REE is slightly less than twice of that of MV-REE, according to MATLAB simulation. 
 \begin{algorithm}[H]
 \caption{Algorithm for C-MV-REE}
 \label{alg:C-MV-REE}
 \begin{algorithmic}[1]
 \renewcommand{\algorithmicrequire}{\textbf{Input:}}
 \REQUIRE $\mathbf{\Delta}^{(v)}$, $\textbf{W}$
 \renewcommand{\algorithmicrequire}{\textbf{Initialization:}}
 \REQUIRE $\textbf{B}^0$, $\eta$, $\sigma$
 \FOR {k = 1 to maximum iteration}
 \STATE $\textbf{B}^k=\textbf{B}^{k-1}+\eta \frac{\partial f}{\partial \textbf{B}}$ as in (\ref{eqn:C-MV-REEsol})
 \STATE Decompose \textbf{B} into $\textbf{U}\mathbf{\Lambda} \textbf{U}^T$ (spectral decomposition)
 \STATE $[\mathbf{\Lambda}_+]_{ij}=\max \{\mathbf{\Lambda}_{ij},0\}$
 \STATE $\textbf{B}^k=\textbf{U}\mathbf{\Lambda_+} \textbf{U}^T$
 \ENDFOR
 \RETURN $\textbf{X}=\textbf{U}\mathbf{\Lambda}^{1/2}$
 \end{algorithmic}
 \end{algorithm}
 
Although all cost functions and derivations in Section \ref{sec:c-mv-feat} and Section \ref{sec:c-mv-ree} pertain to correntropy only, the same optimization schemes work for cost functions using generalized correntropy as well. For C-MV and Ce-MV, the HQ optimization is always applicable because the expectation function always exists. For C-MV-REE, it is obvious that the same gradient ascent method will apply for generalized correntropy with any $\alpha$.
\section{Experiments and result analysis}
Experiments are performed on both simulated data and real-world marine animal data. The main objectives of experimenting with simulated dataset are to study the correntropy loss based multi-view learning algorithms' performance in the existence of noise, compared to algorithms with different cost functions. Multi-view algorithms' superiority over their single view counterparts is also studied. The first real-world data are features from color images, on which the feature-based algorithms (C-MV and Ce-MV) will be tested. The second data are dissimilarity matrices that derives from shape analysis performed on Lidar data, on which the dissimilarity-based algorithm (C-MV-REE) will be tested. C-MV-REE is also suitable for the first dataset because features can be turned into dissimilarity matrices.

\subsection{Experiments on simulated data set}

\subsubsection{UCI handwritten digit classification}
\begin{figure}[t]
\includegraphics[width=10cm,height=4cm]{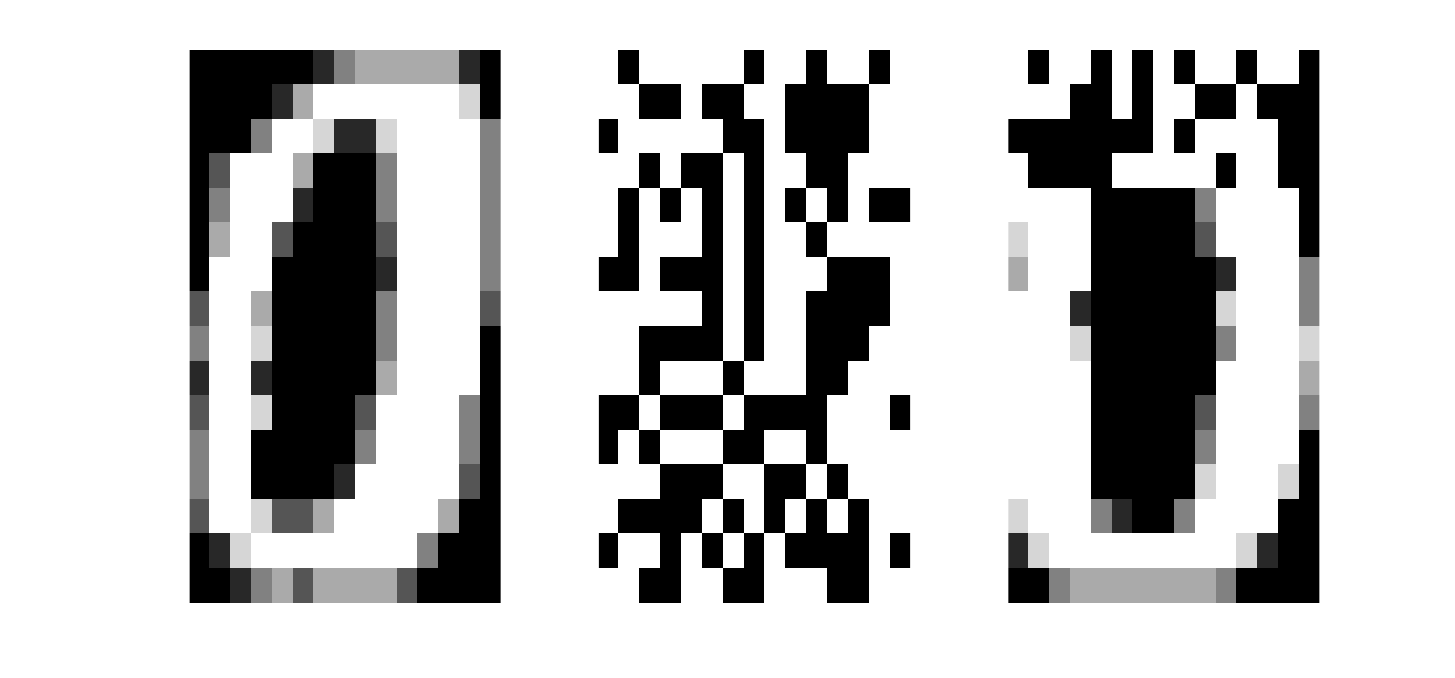}
\centering
\caption{Examples of uncorrupted image, complete salt and pepper noise, and partially noisy image.}
\label{fig:three0}
\end{figure}

The C-MV algorithm and its variant - the Ce-MV algorithm will be tested on the UCI handwritten digit dataset \url{http://archive.ics.uci.edu/ml/datasets/MultipleFeatures}. The dataset has 2000 instances equally divided into 10 classes. Two features used for the $i^{th}$ instance are the image itself stretched into a vector $\textbf{z}^{(1)}_i$ and the Zernike moment $\textbf{z}^{(2)}_i$, whose dimensions are 240 and 47 respectively. Two separate noise conditions are considered for the first feature. In the first condition (Figure \ref{fig:three0} left and center plot), a portion of all images (e.g. 250 images out of 2000) are replaced with salt and pepper noise, whose mean, maximum and minimum pixel values are the same as the rest uncorrupted images. In the second condition (Figure \ref{fig:three0} right plot), a part (e.g. 60 pixels out of 240) of every image are replaced with salt and pepper noise. The other feature, Zernike moments, is left unchanged for both conditions. 

For C-MV, $\sigma=0.5$, while for Ce-MV $\sigma_{(v)}$ are rescaled accordingly. Dimension $d$ of the single-view feature to be learned $\textbf{x}_i$ for both algorithms is set as 60. These parameters are fixed throughout the experiment. The four other methods used for comparison are: (1) using the first feature (image) only; (2) using the second feature (Zernike) only; (3) using the concatenated feature; (4) the $L_2$-MV algorithm, which employs the $L_2$ cost function but is otherwise the same as C-MV. The reason that MISL (which uses Cauchy loss function) is not compared is because its performance is very close to that of C-MV under the noise level in this experiment, according to the analysis in Section \ref{sec:gcloss}. To emphasize the importance of the cost function being bounded, the $L_2$ cost function is used instead. For methods (1)(2)(3), instead of using the features directly, dimensional reduction is applied by feeding two identical feature sets into the same C-MV algorithm to ensure that dimensionality will not be a factor for any difference in performance. The output feature for all 6 methods will be consequently used as input to SVM for classification. 

\begin{figure}[t]
\includegraphics[width=10cm,height=5cm]{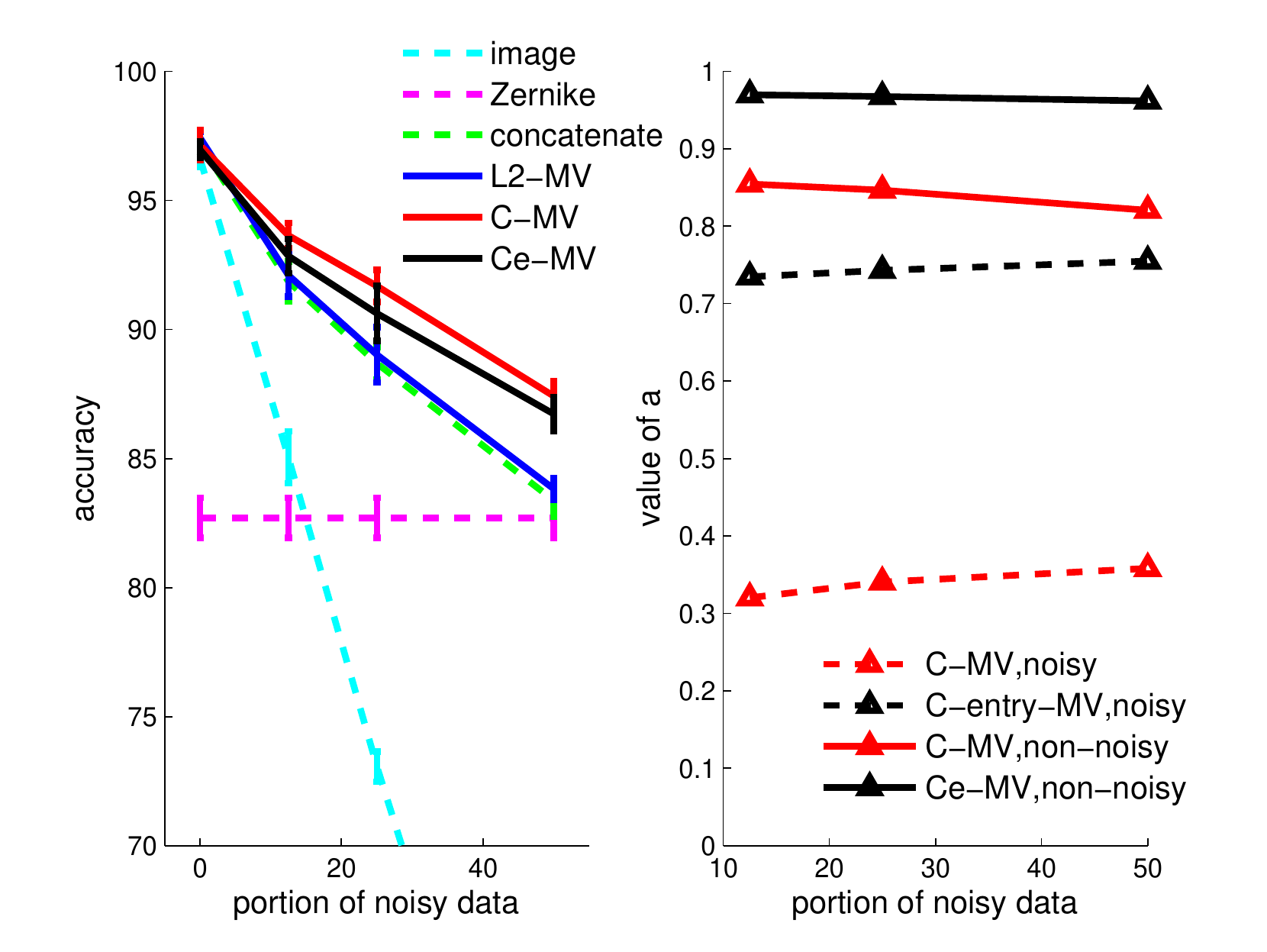}
\centering
\includegraphics[width=10cm,height=5cm]{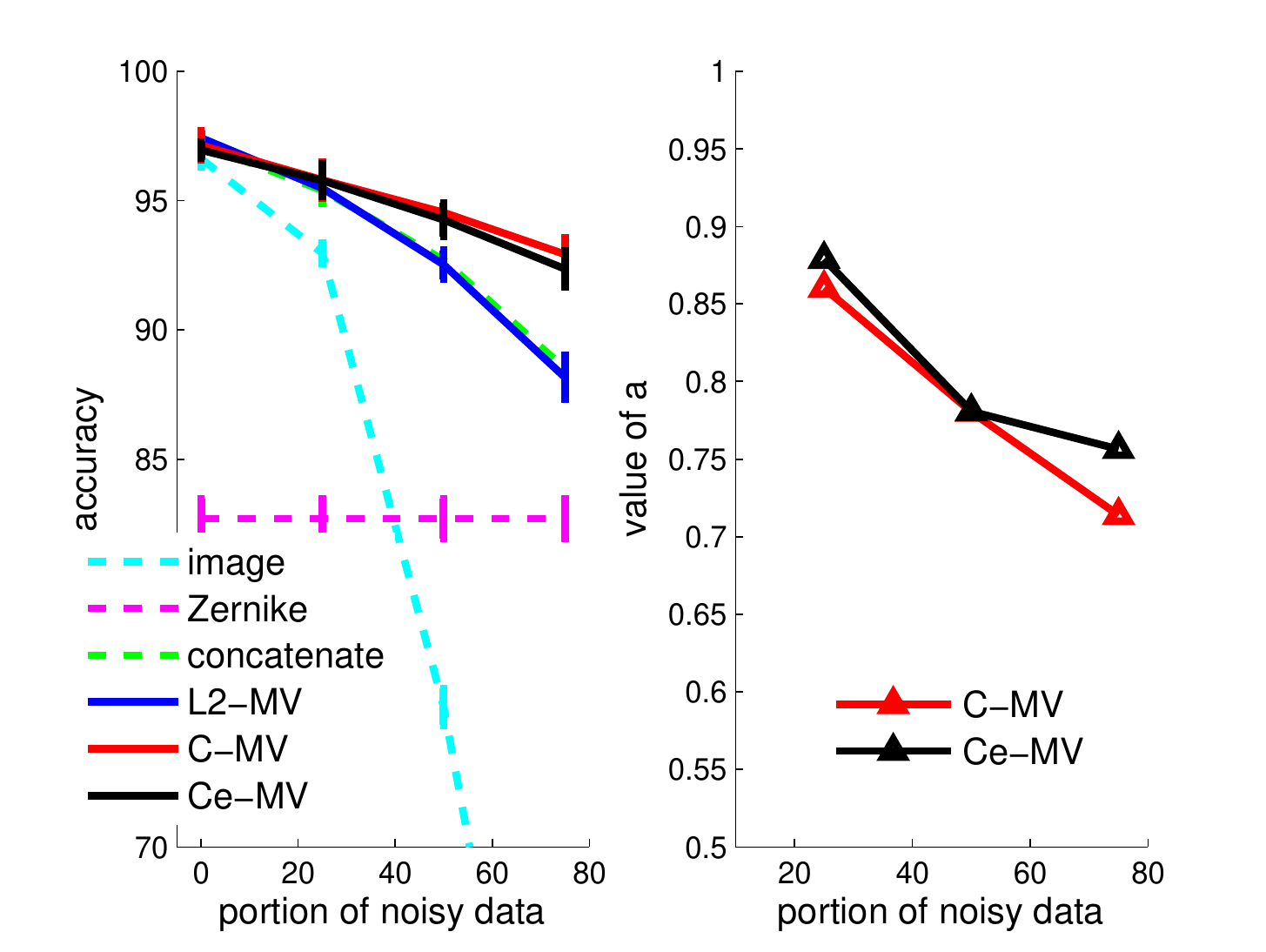}
\centering
\includegraphics[width=10cm,height=5cm]{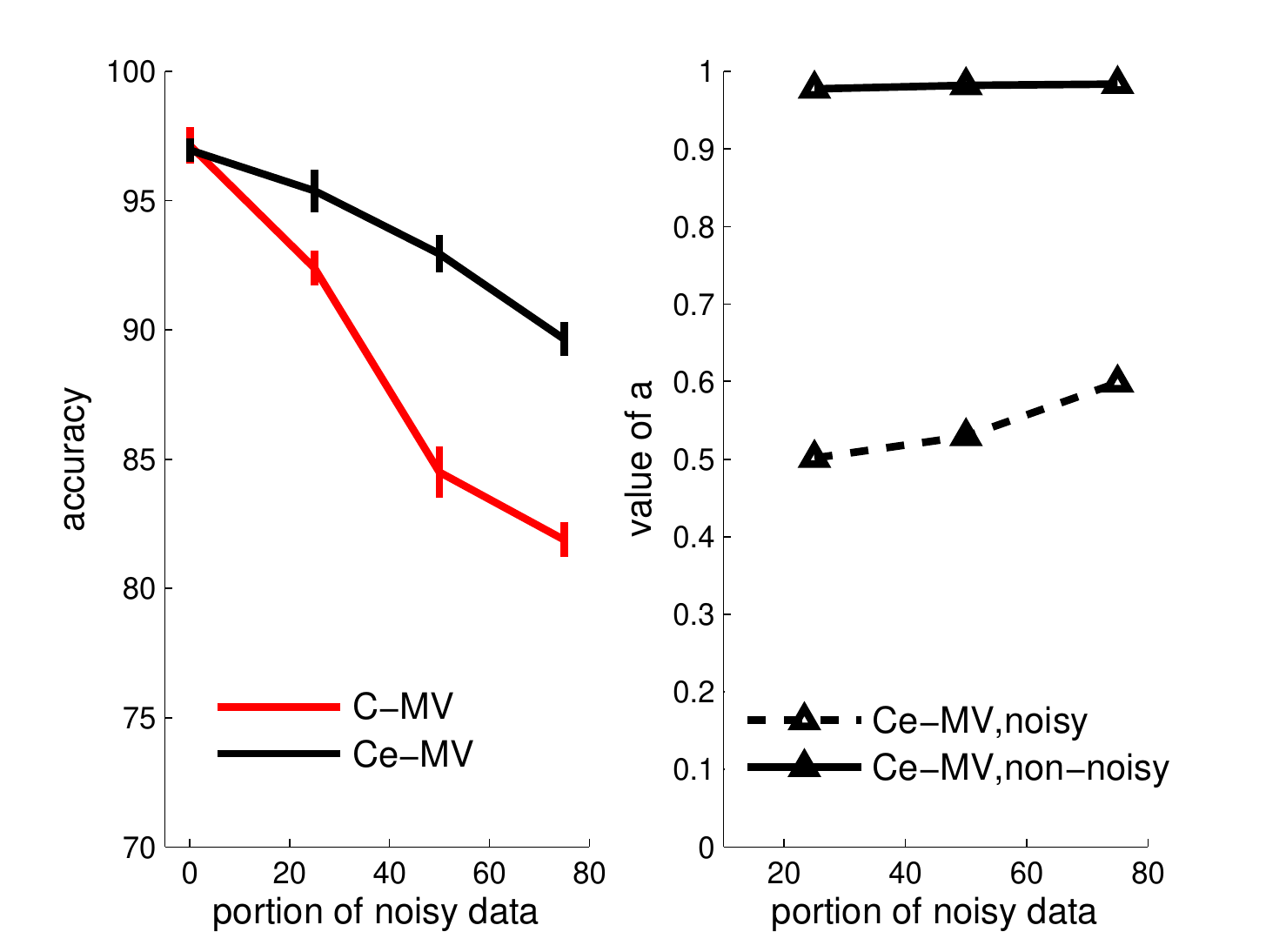}
\centering
\caption{Left column: classification accuracy. Right column: top, center row - mean value of $a^{(1)}_{i}$ (mean across instance) for C-MV and Ce-MV; bottom row - mean value of $a^{(1)}_{j}$ (mean across pixels) for Ce-MV. Note that $a^{(1)}_{j}$ is unavailable for C-MV. The top row corresponds to the first noise condition, where 0\%, 12.5\%, 25\% and 50\% of images are replaced with salt and pepper noise that has the same mean, maximum and minimum pixel values as the rest uncorrupted images. The center row corresponds to the second noise condition, where 0\%, 25\%, 50\% and 75\% of pixels in every image are replaced with the same salt and pepper noise. The bottom row also corresponds to the second noise condition, but the magnitude of noise is 3 times as large as that of the center row. In all cases, mean of $a^{(2)}_{i}$ or $a^{(2)}_{j}$ are very close to 1 (around 0.98) and are not shown in plot.}
\label{fig:UCIresult}
\end{figure}

Results are shown in Figure \ref{fig:UCIresult}. For both noise conditions, applying multi-view learning rather than using a single view will significantly boost classification performance. For the first noise condition (top row), best performances are achieved by C-MV and Ce-MV. This is because C-MV and its entrywise variation are able to distinguish the non-noisy data instances from the noisy ones, which is reflected by $a^{(1)}_i$. The top right plot shows that the mean value of $a^{(1)}_i$ across non-noisy instances is consistently higher than those across noisy instances. Meanwhile, $L_2$-MV assumes the same importance (1) for every instance, while doing feature concatenation instead of multi-view learning will forcefully give the same importance for different views, leading to suboptimal performance. For the second condition with smaller noise level (center row), C-MV and Ce-MV still have better performance. This is because both algorithms will correctly give smaller $a^{(1)}_i$ than $a^{(2)}_i$ across all instances $i$. It can be seen that the more noisy instances there are, the smaller $a^{(1)}_i$ is. The advantage of Ce-MV over C-MV begins to show where noise magnitude is higher (bottom row). As explained in Section \ref{sec:c-mv-feat}, C-MV can weight instances only, not individual features (pixels in this setting). Noise on part of the pixels that is big enough will affect the whole instance. On the other hand, Ce-MV can effectively separate the non-noisy and noisy parts of an instance, giving smaller importance to the noisy pixels and thus enhancing the quality of the learned $\textbf{x}_i$. 

\subsubsection{2-D point set reconstruction}  

\begin{figure}[t]
\includegraphics[width=0.8\linewidth]{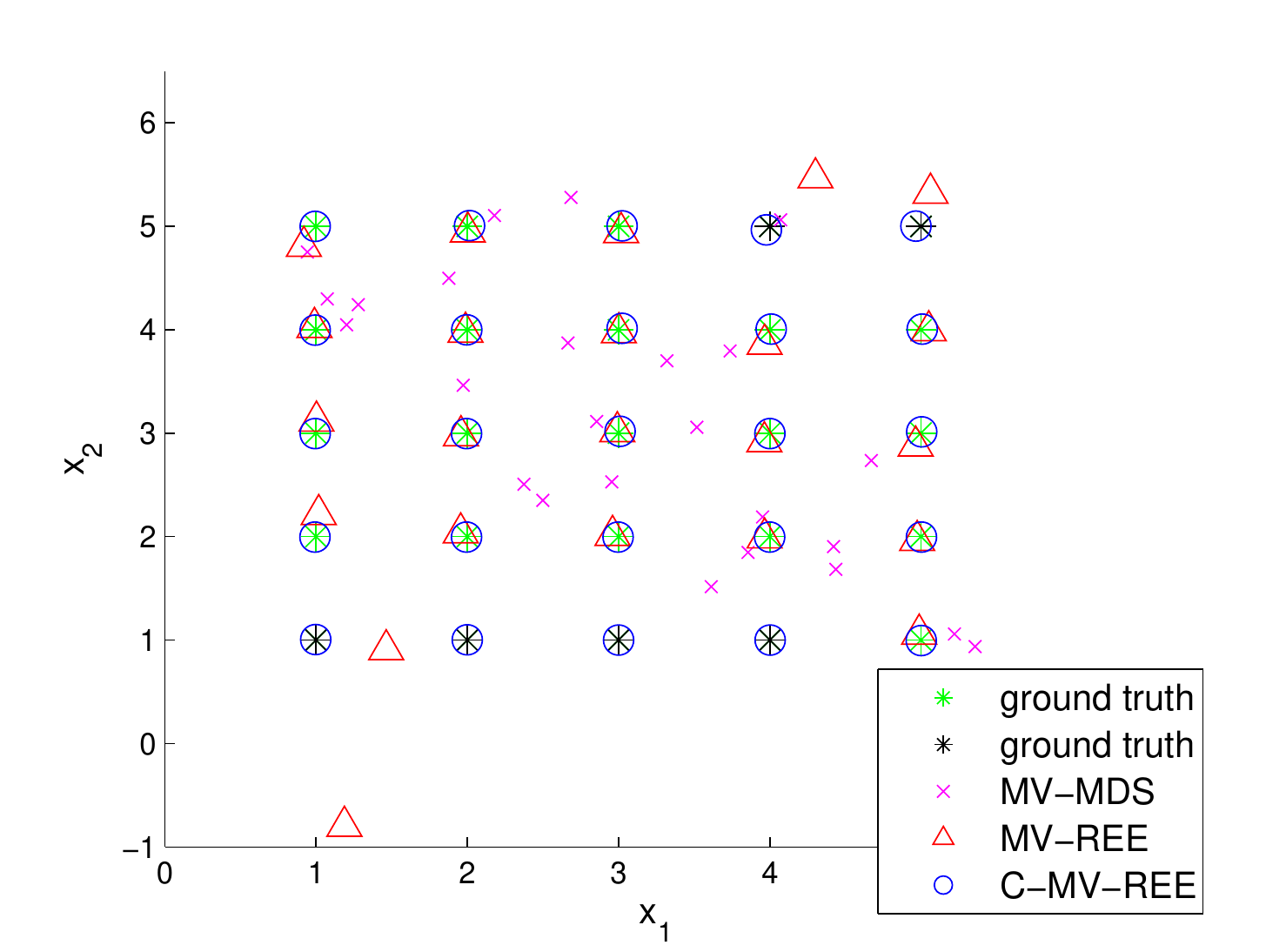}
\centering
\caption{
Comparison of three methods on reconstructing 2-D point set. The original point set is marked with "*", with lighter markers denoting points whose respective distances are uncorrupted by noise, and darker markers denoting otherwise.}
\vspace{-0mm}
\label{fig:conf_a}
\end{figure}

A 2-D point set with $N$=25 points (see Figure \ref{fig:conf_a}) is considered for this experiment. The goal is to recover the original configuration given two different "views" of the point set. The Euclidean distance matrix is computed from the original point set. Noise is then added: for the first view $\mathbf{\Delta}^{(1)}$, all distances pertaining to the first 4 points are corrupted with salt and pepper noise: $ \delta_{i,j}=d_{i,j}+\epsilon_{i,j} \;(i<j)$, where $\epsilon_{i,j}$ has a magnitude of 10. For the second view $\mathbf{\Delta}^{(2)}$, noise of the same type and magnitude corrupts the $24^{th}$ and the $25^{th}$ points. The values of $\delta_{i,j}$ are truncated such that $\delta_{i,j}>0$. The two distance matrices are then symmetrized. 

\begin{figure}[t]
\includegraphics[width=0.75\linewidth]{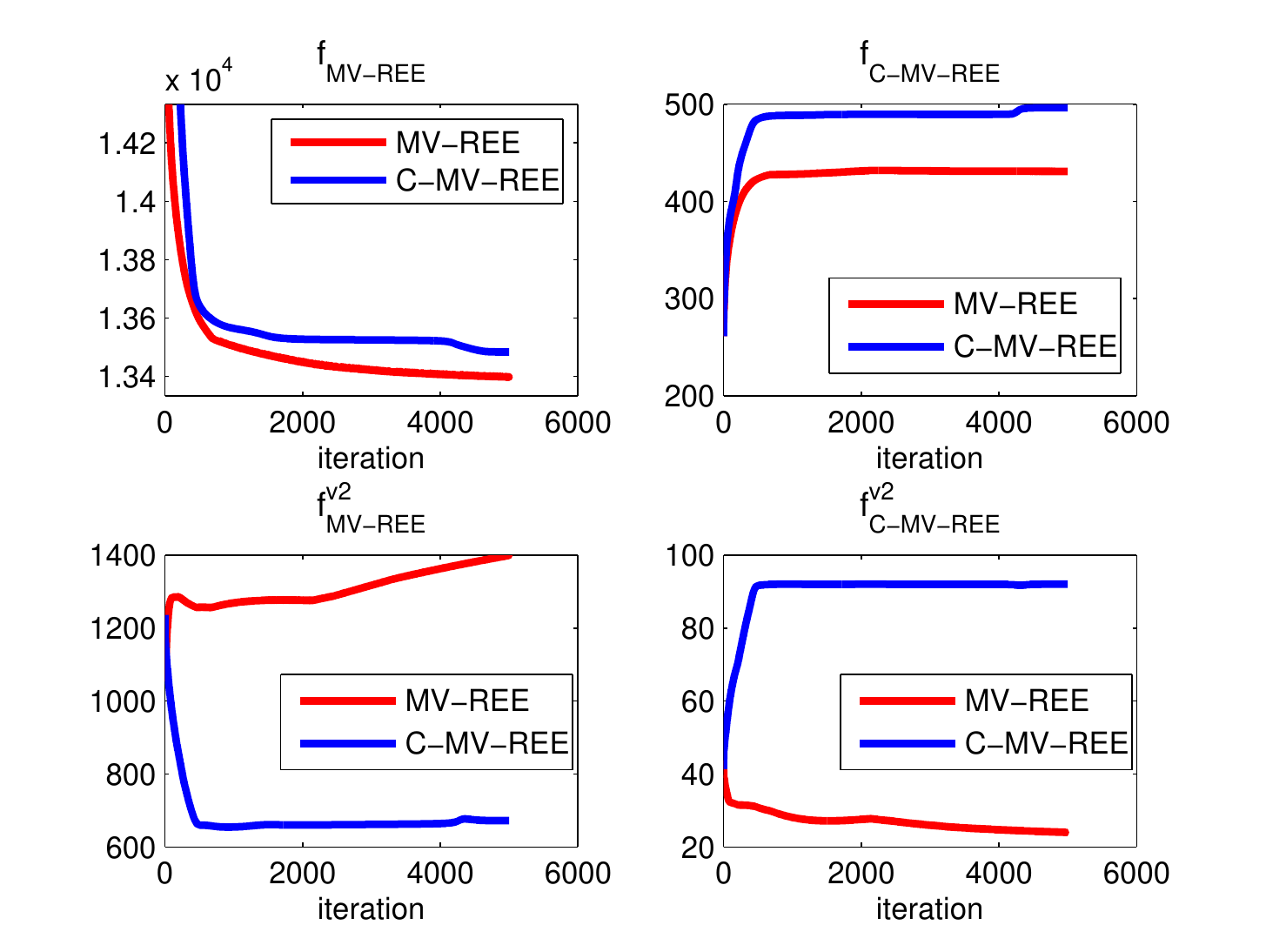}
\centering
\caption{Values of different cost functions dor 2-D point set reconstruction. All red lines represent optimization using MV-REE, and blue lines represent optimization using 
C-MV-REE. The upper left figure and upper right figure are the values of cost functions (\ref{eqn:MV-REE}) and (\ref{eqn:C-MV-REE}), respectively. The lower figures, on the other hand, are values of a "fraction" of the same cost functions: only the second view ($v$=2) is considered, and the range of $i$ is from 1 to 4 rather than 1 to 25 (range of $j$ is still 1 to 25).}
\label{fig:converge}
\end{figure}

Three methods are compared: MV-MDS \cite{Bai16}, MV-REE and C-MV-REE, both of which have been introduced in Section III. For MV-REE, the initial step size $\eta_0$ is chosen to be 0.05. The step size at $i^{th}$ iteration will be $\eta=\eta_0 / \sqrt{i}$ according to the original REE algorithm. For C-MV-REE, $\eta=0.1$ which is fixed throughout the iterations. They are chosen such that convergence speed is approximately the same for both algorithms. They are also small enough to ensure good convergence. Since the median values of $\mathbf{\Delta}^{(1)}$ and $\mathbf{\Delta}^{(2)}$ are 9 and 5 respectively, kernel size $\sigma$ is chosen as 3 for C-MV-REE. From the result (Figure \ref{fig:conf_a}), it can be seen that MV-MDS performs very poorly as it is based on the non-classical MDS framework whose mechanism of preserving the Euclidean distances is weak. On the other hand, C-MV-REE achieved a slightly better performance than MV-REE in recovering points that are uncorrupted by noise, and a much better performance in recovering points that have noisy distances in either views. This can be explained intuitively by the fact that C-MV-REE utilizes well the non-noisy part in one view (e.g. points 1-4 in the second view) that is noisy in the other, which cannot be said for MV-REE. Figure \ref{fig:converge} shows that for both MV-REE and C-MV-REE, the overall cost functions (\ref{eqn:MV-REE}) and (\ref{eqn:C-MV-REE}) are correctly optimized over the iterations; however, as the lower figures suggest C-MV-REE can also correctly optimize the fraction of (\ref{eqn:C-MV-REE}) that corresponds to points 1-4 in the second view, while MV-REE cannot.

\subsubsection{Shape retrieval}  
\begin{figure}[t]
\includegraphics[width=0.6\linewidth]{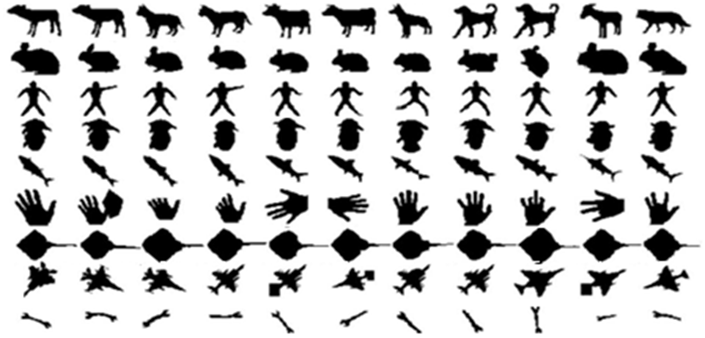}
\centering
\caption{Kimia-99 consists of 9 different classes of objects, with 11 instances in each class.}
\label{fig:kimia99}
\end{figure}

The C-MV-REE algorithm is further tested on the artificial Kimia-99 dataset \cite{Sebastian04}, which has been frequently used as a benchmark for shape retrieval tasks. The dataset is shown in Figure \ref{fig:kimia99}. For every instance, 10 out of 98 most similar instances are found using two algorithms, namely "shape context (SC)" \cite{Belongie02} and "information point set registration (IPSR)" \cite{cao16}. Ideally, the 10 instances should be from the same class as the query instance. The highest total number of correct findings is thus $10*99=990$. No additional noise is added to the data, but the resulting dissimilarity matrices are expected to be noisy due to variation in shapes and imperfection of SC and IPSR algorithms.

\begin{figure}[t]
\includegraphics[width=12cm,height=6cm]{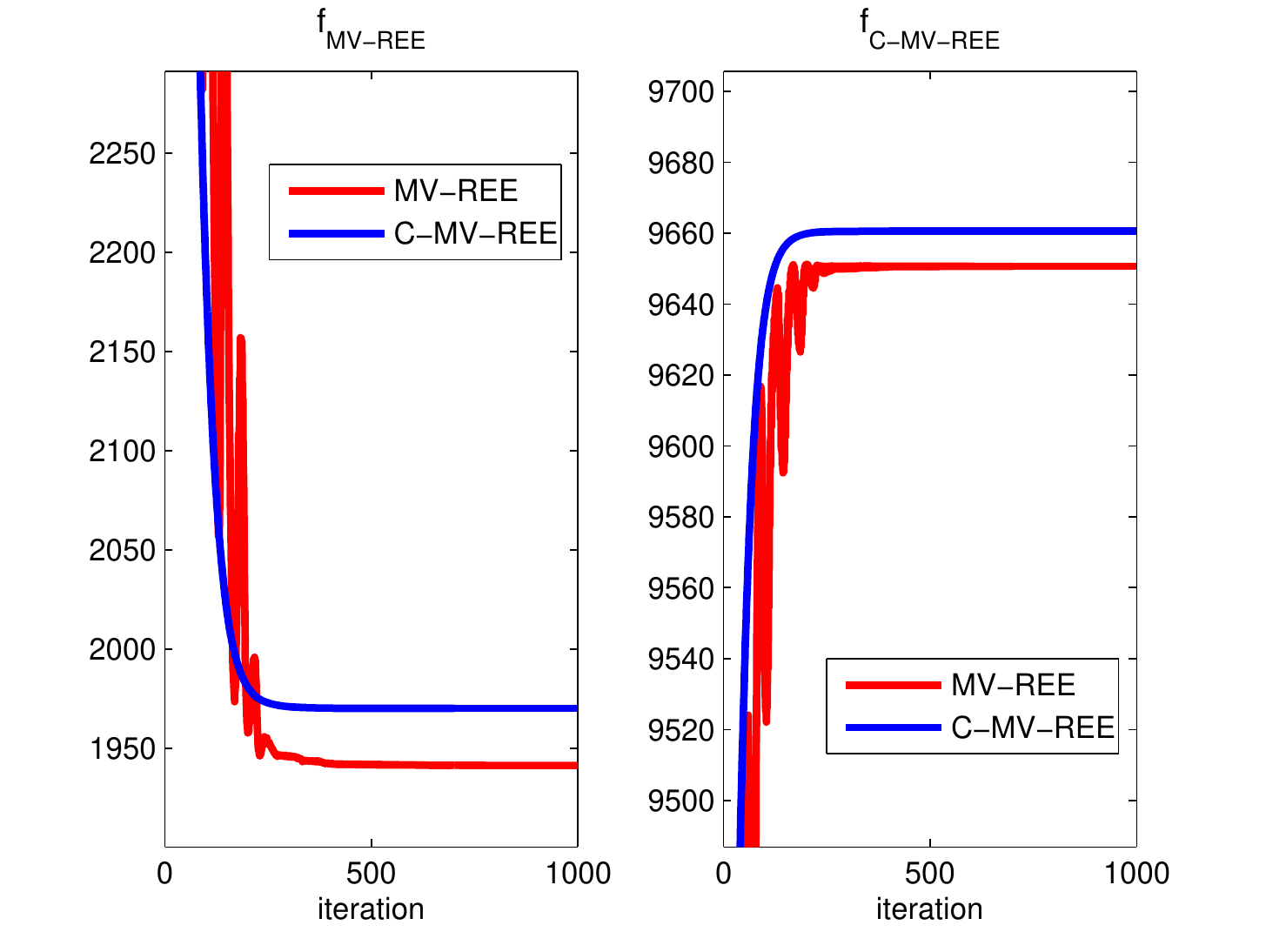}
\centering
\caption{Values of cost functions (\ref{eqn:MV-REE}) and (\ref{eqn:C-MV-REE}) for Kimia-99 dataset. Red lines represent optimization using MV-REE, and blue lines represent optimization using C-MV-REE.}
\label{fig:converge_kimia}
\end{figure}

This experiment will first test the performance of the two algorithm individually. Both SC and IPSR will return 99*99 similarity matrices, which are consequently transformed into proper dissimilarity matrices $\mathbf{\Delta}_{SC}$, $\mathbf{\Delta}_{IPSR}$  (symmetric, main diagonal are zeros). The 10 entries in each row that have the smallest dissimilar values are considered as being from the same class as the query. Next, multi-view learning algorithms are applied to combine $\mathbf{\Delta}_{SC}$ and $\mathbf{\Delta}_{IPSR}$. The following approaches are also tested for comparison purposes: (1) heuristic combination $\mathbf{\Delta}_{heu}$, which is the element-wise square root of the Hadamard product $\mathbf{\Delta}_{SC}\circ\mathbf{\Delta}_{IPSR}$, as in \cite{cao16}; (2) use (single-view) REE only to obtain a new dissimilarity matrix for each of $\mathbf{\Delta}_{SC}$ and $\mathbf{\Delta}_{IPSR}$. Parameter settings for SC and IPSR are same as in \cite{cao16}, except that non-affine transformation for IPSR is not applied in this experiment for its limited effectiveness. Step sizes for MV-REE and C-MV-REE are 0.02 and 0.01 respectively. Since both $\mathbf{\Delta}_{SC}$ and $\mathbf{\Delta}_{IPSR}$ have median values of around 1.9, kernel size $\sigma$ is selected as 1.5. For all approaches that involve REE, the final configuration matrix \textbf{X}' is the first 8 columns of \textbf{X}. Table \ref{tab:kimia_result} summarizes the results. It firstly shows that REE can greatly improve shape retrieval accuracy even if only 1 dissimilarity is considered. This is because REE has the advantages of being a classical MDS algorithm. Meanwhile, the assumption that dissimilarity among instances can be quantified by Euclidean distance is also validated. Secondly, C-MV-REE achieves perfect retrieval result and performs slightly better than MV-REE. Figure \ref{fig:converge_kimia} shows that both algorithms converge well, yet MV-REE has heavier oscillation due to its usage of the subgradient method.

\begin{table}[t!]
\textwidth=7cm \textheight=2cm
\centering
\caption{Kimia-99 shape retrieval results.}
\label{tab:kimia_result}
\resizebox{\textwidth}{\textheight}{  
\begin{tabular}{|c|c|}
\hline
Method & Total correct findings \\ \hline
SC only & 927   \\ \hline
IPSR only & 929  \\ \hline
SC only, REE & 973   \\ \hline
IPSR only, REE & 980   \\ \hline
Heuristic comb. & 948   \\ \hline
MV-REE & 984   \\ \hline
C-MV-REE & 990   \\ \hline
\end{tabular}}
\end{table}

\subsection{Marine animal classification}

\subsubsection{Classification of color images}
\begin{figure}[t]
\includegraphics[width=17cm,height=3.5cm]{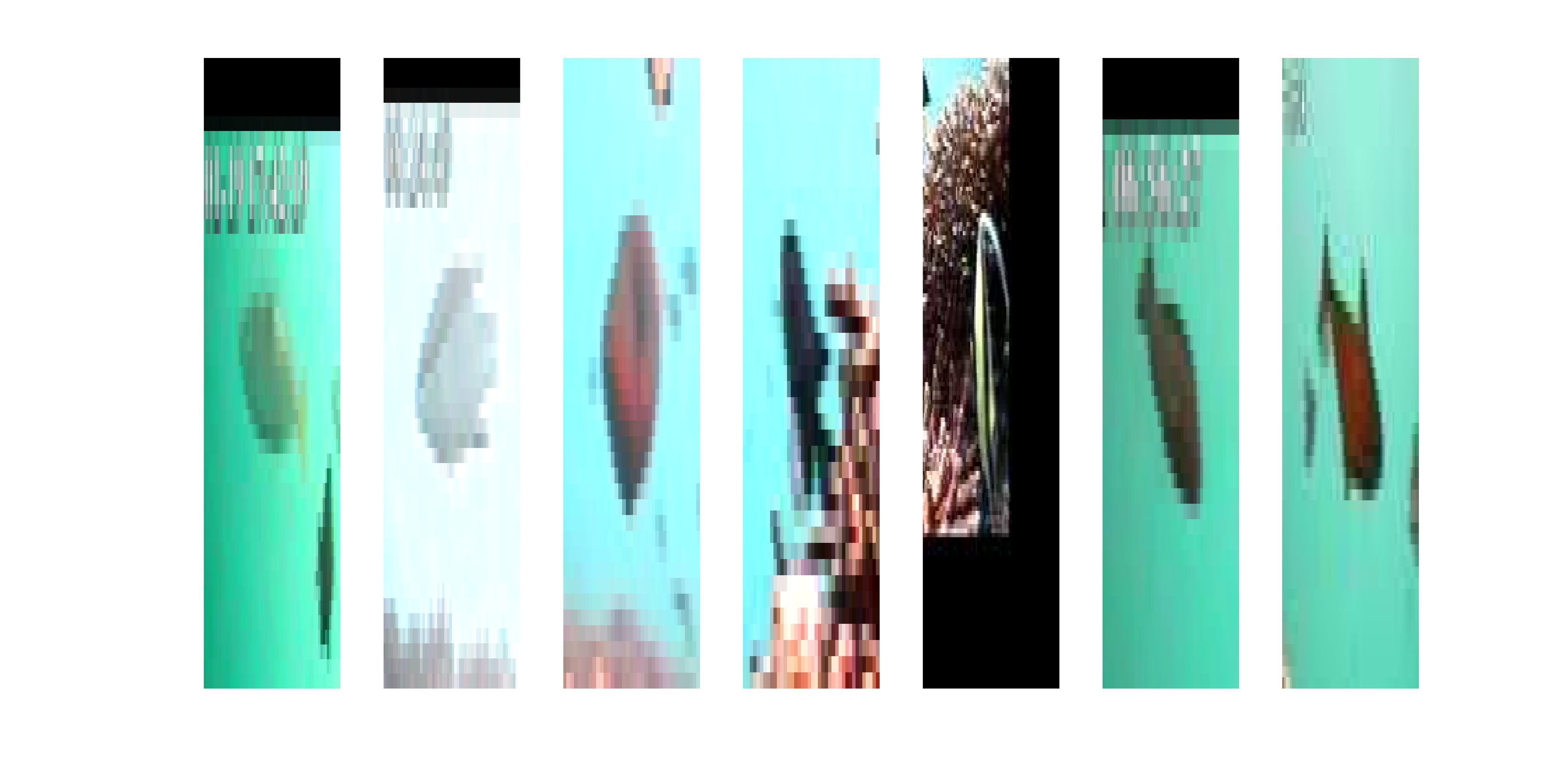}
\centering
\caption{Examples of color images of fish.}
\label{fig:fish7}
\end{figure}

The first marine animal classification experiment is carried out on color images - the Taiwan sea fish dataset (\url{http://sourceforge.net/projects/fish4knowledgesourcecode/}). It has been studied throughout the European Fish4Knowledge project, including works that focus on feature extraction and classification \cite{Spampinato10}\cite{Huang13}. The original dataset has 23 types of fish and over 20,000 fish instances, from which the 7 types with the least number of instances (totaling 237 instances) are used in the experiment (see Figure \ref{fig:fish7}). The two features, namely the CNN (convolutional neural network) feature and the "hand-designed feature", have dimensionalities of 4096 and 2626 respectively. Detailed descriptions of the two features can be seen in \cite{Cao15_2}. 

This experiment will firstly test the feature-based multi-view algorithms. Kernel size $\sigma$ is 1 for C-MV and 0.02 for Ce-MV. Dimension $d$ of \textbf{x} is chosen as 60. Dimensional reduction is also performed when only a single feature (CNN or hand-designed) or the concatenated feature is used. Both SVM and 1-nearest neighbor classifier are tested to evaluate the quality of learned feature more comprehensively. Table \ref{tab:taiwanfeat} compares classification results for 6 methods with different ratios of training and testing data. Reducing the dimension of data from 4096+2626 to a much lower 60 is shown to be beneficial for classification. On the other hand, separately input the CNN and hand features into the multi-view learning algorithms (C-MV or Ce-MV) will result in higher classification accuracy  than concatenating the features.

Furthermore, it will be shown that the dissimilarity-based multi-view algorithm (C-MV-REE) is also a viable option. The Gram matrices computed from features are transformed into dissimilarity matrices first. This experiment will examine whether the REE scheme itself is effective, and whether multi-view learning will outperform heuristic combination of dissimilarity matrix with REE. Kernel size $\sigma$ for C-MV-REE is set as 1 (median values for both $\mathbf{\Delta}_{CNN}$ and $\mathbf{\Delta}_{hand}$ are around 0.55). It turned out that both REE and the correntropy-based multi-view learning are beneficial for improving classification accuracy. Another interesting discovery is that CNN features are not as good as hand-designed feature here, possibly because the generalization ability of a pre-trained CNN is limited on a small dataset which may have heavy bias. 

\begin{table}[t!]
\textwidth=11cm \textheight=4cm
\setlength\extrarowheight{2pt}
\centering
\caption{Taiwan sea fish classification accuracy (feature based). For each method, the first row shows SVM results and the second row shows kNN (k=1) results.}
\label{tab:taiwanfeat}
\resizebox{\textwidth}{\textheight}{  
\begin{tabular}{|l|c|c|c|c|}
\hline
train : test       & 1:4 & 2:3 & 3:2 & 4:1 \\ \hline
CNN only  & 61.9$\pm$11.3\% & 76.6$\pm$7.1\% & 82.5$\pm$8.7\% & 85.8$\pm$12.5\%\\ 
(dimension reduced)  & 66.6$\pm$8.9\% & 77.8$\pm$8.0\% & 80.9$\pm$7.1\% & 81.5$\pm$17.3\%\\ \hline
hand only & 74.9$\pm$8.5\% & 88.1$\pm$6.8\% & 90.6$\pm$6.8\% & 92.8$\pm$7.3\%\\ 
(dimension reduced)  & 78.2$\pm$10.3\% & 84.4$\pm$8.1\% & 87.3$\pm$9.1\% & 89.9$\pm$11.5\%\\ \hline
concatenation & 80.7$\pm$8.7\% & 90.8$\pm$5.5\% & 93.6$\pm$4.8\% & 94.2$\pm$7.6\%\\ 
(no dim. reduction)  & 66.5$\pm$7.7\% & 77.0$\pm$8.4\% & 81.1$\pm$8.3\% & 85.9$\pm$9.6\%\\ \hline
concatenation & 81.6$\pm$10.3\% & 91.4$\pm$6.8\% & \textbf{94.2$\pm$5.7\%} & \textbf{95.5$\pm$7.3\%}\\ 
(dimension reduced)  & 75.6$\pm$12.7\% & 84.4$\pm$9.7\% & 88.9$\pm$7.7\% & 90.9$\pm$9.7\%\\ \hline
C-MV & 82.1$\pm$6.3\% & 90.9$\pm$6.1\% & 93.7$\pm$5.8\% & 94.8$\pm$7.3\%\\ 
  & 82.4$\pm$9.2\% & \textbf{90.7$\pm$6.5\%} & \textbf{93.4$\pm$7.0\%} & \textbf{95.4$\pm$9.2\%}\\ \hline
Ce-MV & \textbf{86.9$\pm$7.3\%} & \textbf{91.9$\pm$6.6\%} & 94.0$\pm$5.8\% & \textbf{95.5$\pm$7.3\%}\\ 
  & \textbf{82.6$\pm$7.3\%} & 88.5$\pm$8.8\% & 92.5$\pm$5.2\% & 94.9$\pm$7.3\%\\ \hline
\end{tabular}}
\end{table}

\begin{table}[t!]
\textwidth=11cm \textheight=4cm
\setlength\extrarowheight{2pt}
\centering
\caption{Taiwan sea fish classification accuracy (dissimilarity matrix based).}
\label{tab:taiwanREE}
\resizebox{\textwidth}{\textheight}{  
\begin{tabular}{|l|c|c|c|c|}
\hline
train : test       & 1:4 & 2:3 & 3:2 & 4:1 \\ \hline
CNN only  & 47.1$\pm$15.4\% & 69.7$\pm$8.0\% & 78.3$\pm$6.4\% & 83.4$\pm$9.9\%\\ 
(with REE)  & 56.9$\pm$4.0\% & 68.5$\pm$6.9\% & 72.7$\pm$8.4\% & 75.4$\pm$13.8\%\\ \hline
hand only & 66.5$\pm$9.8\% & 83.1$\pm$6.8\% & 86.5$\pm$6.8\% & 90.9$\pm$8.1\%\\ 
(with REE)  & 83.1$\pm$6.8\% & 87.0$\pm$8.1\% & 89.7$\pm$7.4\% & 92.1$\pm$8.7\%\\ \hline
Hadamard prod. & 60.3$\pm$11.6\% & 81.6$\pm$9.1\% & 87.7$\pm$6.4\% & 90.7$\pm$7.5\%\\ 
(no REE) & 82.1$\pm$8.0\% & 89.0$\pm$6.8\% & 92.1$\pm$5.9\% & 94.0$\pm$7.4\%\\  \hline
Hadamard prod. & 66.5$\pm$9.5\% & 84.6$\pm$7.8\% & 90.1$\pm$4.9\% & 93.5$\pm$5.0\%\\ 
(with REE) & \textbf{84.5$\pm$7.8\%} & \textbf{90.0$\pm$6.2\%} & 92.5$\pm$6.2\% & \textbf{94.8$\pm$7.6}\%\\  \hline
MV-REE & 67.8$\pm$10.1\% & 84.6$\pm$8.5\% & 88.6$\pm$6.5\% & 90.4$\pm$7.7\%\\ 
 & 79.6$\pm$9.5\% & 87.8$\pm$8.2\% & 91.4$\pm$7.8\% & 94.2$\pm$7.9\%\\  \hline
C-MV-REE & \textbf{70.9$\pm$11.7\%} & \textbf{86.7$\pm$6.0\%} & \textbf{92.7$\pm$4.9\%} & \textbf{95.8$\pm$5.4\%}\\ 
 & 84.2$\pm$7.2\% & 89.5$\pm$7.1\% & \textbf{92.6$\pm$6.3\%} & \textbf{94.8$\pm$7.6\%}\\  \hline

\end{tabular}}
\end{table}

\subsubsection{Classification of Lidar images}
Lidar imagery of fiberglass replicas of three different species of marine animals (amberjack, barracuda and turtle) have been retrieved from the test tank at HBOI. The replicas are mounted on six-degree freedom linear drive such that different poses can be generated. Lidar return of any object firstly undergoes noise reduction and volume backscattering gating. It is then integrated over time, which generates a 2D grayscale image. Contour of the object is then obtained through GrabCut segmentation \cite{Rother04}. The class of any object will be determined by shape analysis, which finds a similarity measure between a pair of contours. Details of Lidar image retrieval, segmentation and preprocessing are described in \cite{Cao16_2}.

There are 38 testing images in total, 22 of which are obtained under clear water condition. The rest are collected under turbid water environment, so it is expected that these 16 objects are heavily blurred, some not even discernible by human standards. The images used as templates (labels known) are 2D projections from different perspectives of 3D models of the three animal species. These 2D images are then processed through a radiative transfer model \cite{Cao16_2}. For each animal specie, 10 2D images of different orientations are generated. The two shape analysis methods used are SC and IDSC. Each method will provide a dissimilarity matrix of size N*N (N=10*3+38=68). Like in the previous experiment, 6 dissimilarity matrix based approaches will be tested. The median value for both $\mathbf{\Delta}_{SC}$, $\mathbf{\Delta}_{IDSC}$ are 3.45 so $\sigma$ is selected as 2 for C-MV-REE. Classification results are average values from 1-NN and SVM classifiers. They are summarized in Table \ref{tab:testtank}. It is evident that the individual method of IDSC performs considerably better than SC in this problem. Using REE on the heuristically combined similarity matrix does not further improve the result. However, the C-MV-REE algorithm still successfully achieves higher classification rate than any of the other methods.  

\begin{figure}[t]
\includegraphics[width=13cm,height=6.5cm]{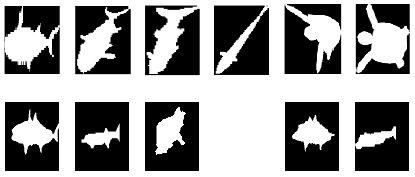}
\centering
\caption{First row: examples of template images. Second row: left: segmented contours of clear water Lidar images; right: contours of turbid water Lidar images.}
\label{fig:fish11}
\end{figure}

\begin{table}[t!]
\textwidth=8cm \textheight=6cm
\caption{Confusion matrices for HBOI test tank data. All results are averaged from the results of SVM and 1NN classifiers. Column I shows the result for all 38 images, while column II is the result for the 22 clear water images. A, B, T stand for amberjack, barracuda and turtle respectively.}
\label{tab:testtank}
\resizebox{\textwidth}{\textheight}{  
\begin{tabular}{|c|c|c|c|c|c|c|c|c|c|}
\w{}&\w{} &\w{}  &\w{I}  &\w{}     &\w{}&\w{}  &\w{II}  &\w{}        \\
\w{}&\w{} &\w{A}     &\w{B}     &\w{T}     &\w{}&\w{A}     &\w{B}     &\w{T}    \\ 
\w{}\\
\w{SC only}&\w{A}&\w{\cm{6.5}}&\w{\cm{8.0}}&\w{\cm{1.5}}& &\w{\cn{3}}&\w{\cn{4.5}}&\w{\cn{0.5}} \\ 
\w{(with REE)}&\w{B}&\w{\cm{1}}&\w{\cm{8}}&\w{\cm{0}}& &\w{\cn{1}}&\w{\cn{5}}&\w{\cn{0}}\\ 
\w{}&\w{T}&\w{\cm{2}}&\w{\cm{5.5}}&\w{\cm{5.5}}& &\w{\cn{1}}&\w{\cn{3.5}}&\w{\cn{3.5}} \\   \w{}\\
\w{IDSC only}&\w{A}&\w{\cm{12.5}}&\w{\cm{0}}&\w{\cm{3.5}}& &\w{\cn{7.5}}&\w{\cn{0}}&\w{\cn{0.5}} \\ 
\w{(with REE)}&\w{B}&\w{\cm{0}}&\w{\cm{9}}&\w{\cm{0}}& &\w{\cn{0}}&\w{\cn{6}}&\w{\cn{0}}\\ 
\w{}&\w{T}&\w{\cm{3}}&\w{\cm{0.5}}&\w{\cm{9.5}}& &\w{\cn{2}}&\w{\cn{0.5}}&\w{\cn{5.5}} \\   \w{}\\
\w{Hadamard prod.}&\w{A}&\w{\cm{12.5}}&\w{\cm{0.5}}&\w{\cm{3}}& &\w{\cn{6}}&\w{\cn{0.5}}&\w{\cn{1.5}} \\ 
\w{no REE}&\w{B}&\w{\cm{6}}&\w{\cm{3}}&\w{\cm{0}}& &\w{\cn{3.5}}&\w{\cn{2.5}}&\w{\cn{0}}\\ 
\w{}&\w{T}&\w{\cm{7}}&\w{\cm{0}}&\w{\cm{6}}& &\w{\cn{4}}&\w{\cn{0}}&\w{\cn{4}} \\   \w{}\\
\w{Hadamard prod.}&\w{A}&\w{\cm{11.5}}&\w{\cm{0}}&\w{\cm{4.5}}& &\w{\cn{6.5}}&\w{\cn{0}}&\w{\cn{1.5}} \\ 
\w{(with REE)}&\w{B}&\w{\cm{1.5}}&\w{\cm{7.5}}&\w{\cm{0}}& &\w{\cn{0.5}}&\w{\cn{5.5}}&\w{\cn{0}}\\ 
\w{}&\w{T}&\w{\cm{2}}&\w{\cm{0}}&\w{\cm{11}}& &\w{\cn{1}}&\w{\cn{0}}&\w{\cn{7}} \\  
\w{}\\
\w{}&\w{A}&\w{\cm{13}}&\w{\cm{0}}&\w{\cm{3}}& &\w{\cn{7.5}}&\w{\cn{0}}&\w{\cn{0.5}} \\ 
\w{MV-REE}&\w{B}&\w{\cm{1}}&\w{\cm{8}}&\w{\cm{0}}& &\w{\cn{0.5}}&\w{\cn{5.5}}&\w{\cn{0}}\\ 
\w{}&\w{T}&\w{\cm{2}}&\w{\cm{0}}&\w{\cm{11}}& &\w{\cn{1}}&\w{\cn{0}}&\w{\cn{7}} \\     
\w{}\\
\w{}&\w{A}&\w{\cm{13}}&\w{\cm{0}}&\w{\cm{3}}& &\w{\cn{7}}&\w{\cn{0}}&\w{\cn{1}} \\ 
\w{C-MV-REE}&\w{B}&\w{\cm{1}}&\w{\cm{8}}&\w{\cm{0}}& &\w{\cn{0}}&\w{\cn{6}}&\w{\cn{0}}\\ 
\w{}&\w{T}&\w{\cm{1.5}}&\w{\cm{0}}&\w{\cm{11.5}}& &\w{\cn{0.5}}&\w{\cn{0}}&\w{\cn{7.5}} \\
\end{tabular}}
\end{table}

\section{Conclusion}
The correntropy loss based multi-view learning algorithms are developed for both cases of features (C-MV, Ce-MV) and dissimilarity matrices (C-MV-REE). In the presence of artificial noise, methods based on correntropy loss are much more robust than their counterparts that apply $L_1$ or $L_2$ loss as cost functions. Multi-view learning performs also better than using single view only or concatenated feature/heuristically combined dissimilarity matrix. The developed algorithms are successfully applied to real-world marine animal classification problems, which is valuable for future studies in which new features or dissimilarity measures are involved. Since the optimization techniques for both C-MV and C-MV-REE can accommodate generalized correntropy loss, it might be interesting in future work to probe into the possibility of employing GC-loss in multi-view learning.



\bibliographystyle{IEEEtran}
\bibliography{joebib}{}
\bibliographystyle{unsrt}

\begin{IEEEbiography}[{\includegraphics[width=1in,height=1.25in,clip,keepaspectratio]{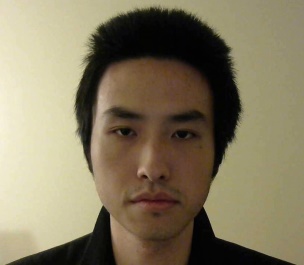}}] {Zheng Cao}
is currently a PhD student in the Computational NeuroEngineering Laboratory (CNEL) at the University of Florida. He received his BS degree from Nanjing University of Science and Technology in 2010, and MS degree from University of Wyoming in 2012. His current research interests are machine learning and computer vision, with applications in marine animal detection and classification. He has previously worked on system identification and kernel adaptive filtering.
\end{IEEEbiography}

\begin{IEEEbiography}[{\includegraphics[width=1in,height=1.25in,clip,keepaspectratio]{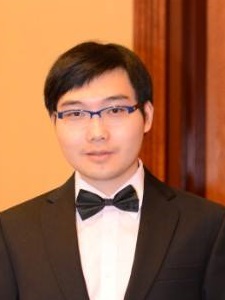}}] {Shujian Yu}
received the B.S. (Hons). degree from the School of Electronic Information and Communications, Huazhong University of Science and Technology, Wuhan, China, in 2013. He is currently pursuing the Ph.D. degree with the Department of Electrical and Computer Engineering, University of Florida, Gainesville, FL, USA. His research interests include machine learning for signal processing, image analysis, and understanding.
\end{IEEEbiography}

\begin{IEEEbiography}[{\includegraphics[width=1in,height=1.25in,clip,keepaspectratio]{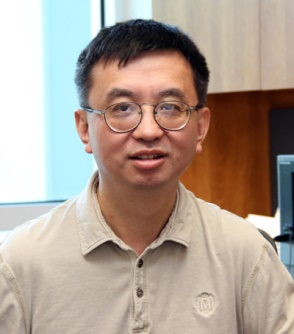}}] {Bing Ouyang}
(S'02-M'06) received the Ph.D. degree in electrical engineering from Southern Methodist University, Dallas, TX, USA, in 2007. He joined the Ocean Visibility and Optics Laboratory, Harbor Branch
Oceanographic Institute (HBOI), Florida Atlantic University (FAU), Fort Pierce, FL, USA, in 2009. Before joining HBOI, he was with Texas Instruments, Inc. (TI), Dallas, TX, USA. From 2003 to 2009, he was an Algorithm Engineer with the DLP ASIC algorithm team, with the primary focus in developing front end algorithms for the video processing ASIC. He holds four U.S. patents in the area of analog video and graphics format detection. His current research interests include underwater computer vision, novel underwater electro-optical system design, underwater LIDAR imaging enhancement, pattern recognition, and analysis for sensor time series data. Dr. Ouyang is a recipient of the 2013 Young Investigator Research Program award. He was peer elected to the Member of Technical Staff while working at TI.
\end{IEEEbiography}

\begin{IEEEbiography}[{\includegraphics[width=1in,height=1.25in,clip,keepaspectratio]{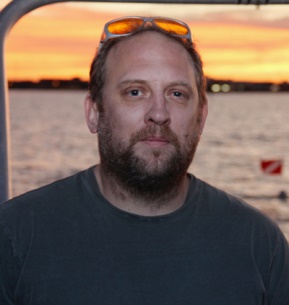}}] {Fraser Dalgleish}
received the Ph.D. degree in ocean engineering from Cranfield University, Cranfield, Bedford, U.K., in 2004. He directs the Ocean Visibility and Optics Laboratory, Harbor Branch Oceanographic Institute (HBOI), Florida Atlantic University (FAU), Fort Pierce, FL, USA. His research emphasis is on undersea optical sensor development, both for remote and in situ environmental measurements and to improve sensing and communications capabilities as an enabling technology for multivehicle imaging and sensing operations. Recent development activities in collaboration with government and industry have focused on new laser instrumentation, simulation tools, and detection approaches for ocean sensing and monitoring applications. Projects within academia involve collaboration with scientists in sensor package development and ocean observatories for water quality monitoring and imaging missions.
\end{IEEEbiography}

\begin{IEEEbiography}[{\includegraphics[width=1in,height=1.25in,clip,keepaspectratio]{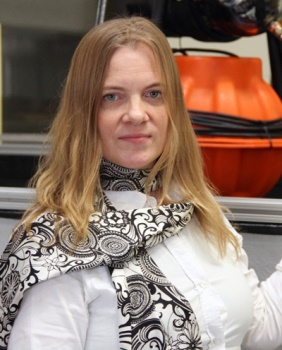}}] {Anni Vuorenkoski}
received the Ph.D. degree in mechanical engineering from Cranfield University, Cranfield, Bedford, U.K., in 2004. She is with the Ocean Visibility and Optics Laboratory, Harbor Branch Oceanographic Institute (HBOI), Florida Atlantic University (FAU), Fort Pierce, FL, USA. Her current research concentrates on the characterization of water column and benthic features by polarimetry, fluorometry, angularly resolved scattering techniques, and time-resolved short-pulse methods, as well as on the experimental validation of computational radiative transfer models. Her past research activities have included adoption of optical, laser-based techniques to study the properties of turbid media and flows. She has also been involved in the development and application of Monte-Carlo–based computational methods to simulate the effects of multiple scattering in aerosol laser imaging.
\end{IEEEbiography}

\begin{IEEEbiography}[{\includegraphics[width=1in,height=1.25in,clip,keepaspectratio]{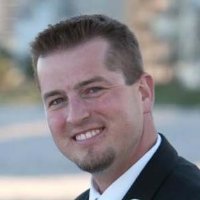}}] {Gabriel Alsenas}
received his B.S. and M.S. degree in Ocean and Systems Engineering from Florida Atlantic University (FAU), in 2005 and 2007 respectively. He has been the program manager at FAU Southeast National Marine Renewable Energy Center for 7 years.
\end{IEEEbiography}

\begin{IEEEbiography}[{\includegraphics[width=1in,height=1.25in,clip,keepaspectratio]{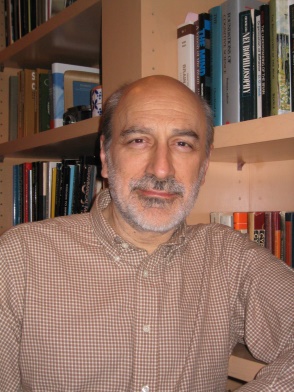}}] {Jose C. Principe}
(F'00) is a Distinguished Professor of Electrical and Computer Engineering and Biomedical Engineering at the University of Florida. He is also the BellSouth Professor and Founding Director of Computational NeuroEngineering Laboratory (CNEL), University of Florida. His primary research interests are advanced signal processing with information theoretic criteria (entropy and mutual information), adaptive models in the reproducing kernel Hilbert spaces (RKHS) and the application of these advanced algorithms in Brain Machine Interfaces (BMI). Dr. Principe is a Fellow of the IEEE, ABME and AIBME. He is the past Editor-in Chief of the IEEE Transactions on Biomedical Engineering, past Chair of the Technical Committee on Neural Networks of the IEEE Signal Processing Society and past President of the International Neural Network Society. He received the IEEE EMBS Career Award, and the IEEE Neural Network Pioneer Award. He has more than 600 publications and 30 patents.
\end{IEEEbiography}

\end{document}